# Ten Hard Problems in Artificial Intelligence We Must Get Right

GAVIN LEECH*, University of Bristol and Arb Research,

SIMSON GARFINKEL*, BasisTech, LLC,

MISHA YAGUDIN, ALEXANDER BRIAND, and ALEKSANDR ZHURAVLEV, Arb Research,

We explore the AI2050 "hard problems" that block the promise of AI and cause AI risks: (1) developing general capabilities of the systems; (2) assuring the performance of AI systems and their training processes; (3) aligning system goals with human goals; (4) enabling great applications of AI in real life; (5) addressing economic disruptions; (6) ensuring the participation of all; (7) at the same time ensuring socially responsible deployment; (8) addressing any geopolitical disruptions that AI causes; (9) promoting sound governance of the technology; and (10) managing the philosophical disruptions for humans living in the age of AI. For each problem, we outline the area, identify significant recent work, and suggest ways forward. *Note: this paper reviews literature through January 2023.*

## 1 Introduction

Artificial intelligence has begun to revolutionize science and engineering through the automated discovery of new drugs, building materials, and even techniques for fusion energy (see HP#4). At the same time, current generative AI systems can create targeted disinformation and automate the production of hate speech—and these are only two examples of how the benefits and dangers of AI have progressed from hypotheticals to actual issues we must address in the near term.

In response to such opportunities and risks, the AI2050 program was initiated, drawing on previous research and numerous conversations with experts.[1] This paper reviews the AI2050 list of hard problems that AI researchers must "get right" for socially positive outcomes to result by the year 2050 [270]. It attempts to bridge all the fields and factions that concern themselves with AI.

This paper explores each of the ten Hard Problems (HPs). We begin with the recent history of AI in Section 2. Section 3 expands each HP into a research agenda, identifies significant work published in it between 2017 and 2022, and identifies the kinds of research that will be needed in coming decades. A similar exercise was undertaken in Gil and Selman [203]. Calling these problems "hard" might connote that they are unsolvable: "wicked problems" [479]. We address this question in Section 4, concluding that there is hope for partial solutions to suitably sharpened versions of the problems. Section 5 concludes with our outlook for the future. The electronic appendix provides additional resources for each HP and presents our methodology. Throughout this paper, our analysis is guided by AI2050's motivating question: "It's 2050. AI has turned out to be hugely beneficial to society. What happened? What are the most important problems we solved and the opportunities and possibilities we realized to ensure this outcome?" Accordingly, the headline text of each section is subjunctive, conditioned on our collective success in resolving each problem.

---

*Corresponding authors. Contribution statement: Simson Garfinkel is the author of the introduction, the discussion of wicked problems and the conclusion. Simson Garfinkel and Gavin Leech jointly participated in the historical section. The analysis of each specific hard problem is the work of Gavin Leech, Aleksandr Zhuravlev, Misha Yagudin, and Alexander Briand, with editing by Simson Garfinkel.

[1]AI2050 is a $125 million, five-year commitment by Eric and Wendy Schmidt to "support exceptional people working on key opportunities and hard problems that are critical to get right for society to benefit from AI." [497]

---

Authors' addresses: Gavin Leech, g.leech@bristol.ac.uk, University of Bristol and Arb Research, Bristol, ; Simson Garfinkel, simsong@basistech.com, BasisTech, LLC, ; Misha Yagudin; Alexander Briand; Aleksandr Zhuravlev, Arb Research, Prague,

## 2 Background

> "Oh emperor, my wishes are simple. I only wish for this. Give me one grain of rice for the first square of the chessboard, two grains for the next square, four for the next, eight for the next and so on for all 64 squares, with each square having double the number of grains as the square before."
> — *The Rice and Chessboard legend* [124]

From the coining of the term "artificial intelligence" onward [375], progress in AI has been coupled to improvements in computing power and data storage [504]. For decades apparent progress was slow—or even invisible to those following other areas of computer science, where progress was in line with the optimistic predictions of Moore's Law [394] in the 1990s. It appears, to adopt AI pioneer Ray Kurzweil's metaphor [315], that we are now in the "second half of the chessboard," a period in which the acceleration of AI capabilities has become impossible to ignore. There is a good chance that many of the specific claims in this paper about AI limitations will be invalid within one or two years.

Big improvements in AI rely on scaling up compute [46], including (at the top end) by harnessing tens of thousands of machines and millions of cores towards the same problem, and using distributed storage to make proportionally more data accessible at speed. This in turn is dependent on breakthroughs in parallelized training [406, 258, 442], which make it possible to apply thousands of machines to the single problem of tuning the billions or trillions of weights in a large-scale neural network.

As these constraints eased, the last decade saw qualitative changes in AI capabilities, with progress on some of the field's earliest, grandest aspirations [176, 363]. Systems display above-human-level perception on some tasks [304], as well as some multi-task [460, 79, 428, 472] and multi-modal ability [462, 602, 428].[2] This capability is most impressive (and practically usable) in domains involving image, text, or audio data, as well as discrete spaces like game-playing, biological codes, and mathematics, in which we see human-level capability or beyond [217, 487, 544, 499, 277, 343]. In an unprecedented turn, these same systems often present users with powerful natural language user interfaces, with passable performance on *arbitrary* queries [79, 472]. And, most recently, systems that exhibit human-like creativity have sparked widespread interest, as in the case of diffusion models for generating high-resolution, high-quality custom images and videos from given *arbitrary* prompts (see DALL-E, Stable Diffusion, and Phenaki) [467, 484, 574].

*Shortfalls & costs*   But the grandest promises remain unfulfilled. Self-driving cars continue to hover around the bar of practicality after ten years of testing on public roads [522, 166]. Despite IBM's claims that its Watson system would revolutionize healthcare [239], AI for healthcare is only slowly penetrating clinical practice—reflecting "issues with reliability, mixed impacts on workflows, poor user-friendliness and lack of adeptness with local contexts", as well as "limited data availability, trust and evidence of cost-effectiveness" [105], especially in the low and middle-income countries where AI would have the greatest impact due to the sometimes drastic shortage of practitioners. Within AI capabilities, most systems completely lack persistence (saved states and coherent output across sessions), range (long time horizons and hierarchical plans), and continual learning (retaining relevant past information while updating to include new information) [571].

Further, the benefits of AI are not equally experienced. In recent years it has been widely reported that AI systems deployed for applications assessing human beings for credit, employment, and release from prison seem to replicate pre-existing racial biases and stereotypes [402, 463, 475]. Another example is provided by commercially deployed object recognition systems, which seem to do a better job of recognizing objects in wealthy households than those in households at the other end of the economic spectrum [575].

---

[2]A *multi-task* system is one which can solve several unrelated tasks with only minimal further learning 'in-context' [225]. A *multi-modal* system is one that can solve complex tasks using inputs from several 'senses' at once.

Finally, the impressive capabilities of the recent AI revolution have come at significant cost: as of writing, global spending on AI research and development (R&D) is now in the hundreds of billions USD per year[3], with the cost of a single model training run sometimes exceeding $10 million of compute alone [506]. In response, the computer scientist Christopher Manning quipped "you often use the slogan 'AI is the new electricity'... discussing this trend of bigger and bigger models, I've been turning it around, and saying that electricity is the new AI" [413]. However, once the model has been trained (and the upfront cost for the inference infrastructure paid), models have a very low cost per query [145]—and this combination of large upfront capital requirements and economies of scale could easily lead to AI being dominated by a small number of well-funded organizations, blocking entrants to research and markets [14].

These trends, if they continue, could significantly limit the benefits of AI in the coming decades.

## 2.1 Deep Learning Liftoff (2010–2016)

Recent successes in AI are largely due to deep learning (DL), belatedly vindicating the old connectionist research program [370]. In DL, a neural network learns a hierarchical representation of data from examples—including so-called "unstructured" high-dimensional examples (such as arrays of pixels, audio waveforms, and text documents)[4] [56].

Neural networks first appeared in the 1960s [587]; why did they take fifty years to become dominant?

*Inputs: compute and data*    The reductive answer (which completely abstracts out the role of research) is simply that it is computationally expensive to train networks, and the cost of relevant computations decreased by a factor of trillions over this period [265, 264, 507]. At the same time, the internet dramatically reduced the cost of both data acquisition and labeling for key tasks like image recognition and machine translation [53].

The computer scientist Sara Hooker notes that an idea's success in the computational sciences is not just a function of the idea's quality, but also of the hardware and software that enable the idea to demonstrate a practical advantage [249]. This angle explains the decades of no progress in deep learning as "due in large part to incompatible hardware... The inability to parallelize on CPUs meant matrix multiplies quickly exhausted memory bandwidth, and it simply wasn't possible to train deep neural networks with multiple layers." From this perspective, DL won a "hardware lottery" in the 2010s after billions of R&D dollars were spent developing graphical processing units (GPUs), which use matrix multiplications to create realistic 3D graphics. It so happened that GPUs optimized for consumer entertainment also worked for accelerating neural network training [249].

This is obviously an incomplete explanation: Figure 1 describes many other innovations involved in the 2010–2012 computer vision breakthrough that heralded deep learning [108, 310]. That breakthrough, in turn, stimulated the *circa* 2016 influx of researchers and capital that has driven subsequent progress [14]. Even so, the unprecedentedly low cost of computation and data is certainly a key condition of the present paradigm, since deep learning is so far one of few methods capable of using data and compute at the "warehouse computing" scale.

*Correcting misconceptions*    A series of popular errors among researchers constituted another blocker. Over the decades, the received view was that multilayer networks were too hard to train [58] because of instability (vanishing or exploding gradients [416]) or local minima [75, 528, 185]. Additionally, a misapplication of statistical learning theory suggested that overfitting was unavoidable for complex models like large neural networks [56, 603], and the field thus trapped itself in a local maximum of low-parameter models (the "underparameterized" regime) [431, 404]. Sociological factors also held back DL, as with the counter-advocacy of the leading AI scientist

---

[3]Consider the $95.99bn of world private investment in AI in 2023, plus $11.7bn of public spending from the US alone [**index2024**].

[4]The term "unstructured" is an artefact from when most machine learning was applied to tabular data, while images, sounds, and text could not be analyzed without significant preprocessing. In fact these modalities have rich structure, in the form of dependencies in time, space, and more. Modern deep learning techniques excel because they identify and exploit such structures. [242, 492].

Marvin Minsky [386]. With his collaborator Seymour Papert, Minsky proved that shallow neural networks lacked the expressiveness for complex tasks, convincing two generations of researchers to largely avoid them [425]. On one occasion, Minsky asked a presenter "How can an intelligent young man like you waste your time with something like [neural networks]? ... This is an idea with no future." [378].

*The "Common Task Framework"* [154]. ML research is often structured as a competition to perform a fixed benchmark task, with "held-out" test data (ideally not seen by competitors) used to objectively test whether systems generalize beyond the data used to train them. For instance, success on the ImageNet visual classification benchmark was a watershed moment for the popularization of deep neural networks [144, 310]. The degree to which performance on these simple curated datasets generalizes to real-world data is controversial, but it is now accepted that neural networks can learn generally useful representations from training on simple tasks (rather than learning only spurious or highly task-specific features) [139, 56, 79, 344].

*Accumulating training tricks* Much of the progress of the last decade is due to engineering optimizations: training stabilizers, *ad-hoc* algorithmic speedups, and dozens of tweaks to architectures. Together with improvements in the availability of computation and data, these produced a qualitative leap in capabilities thanks to an increased ability to search the vast model space [415]. Figures 1 and 3 describe some of these tricks; other crucial examples include skip connections [433], batch normalization [266], better weight initializations [206, 232], gradient clipping [443], and a plethora of data augmentation methods [195].

Which of these were crucial? A concise enumeration for the ImageNet benchmark is given by Arora and Zhang [28]; taking the best practices of 2011 as background, they give a complete specification for training a 95% accuracy model—including all hyperparameter values—in just 1032 bits. The *minimal* innovations were batch normalization, skip connections, and residual blocks.

*Representation learning and end-to-end training* Previous ML systems involved a pre-processing pipeline of modules which transformed the input data into a form suitable for the final predictor. This was most intense in natural language processing (NLP): before 2018, feature pipelines were elaborate, involving sentence segmentation, word tokenization, lemmatization, stop word removal, part-of-speech tagging, dependency parsing, and constituency phrasing [548]. Many of these modules had to be hand-coded [178] or trained independently. In end-to-end training, conversely, one neural network represents the entire target system on its own, taking raw (or nearly raw) inputs [205]; gradients are backpropagated from the output *end* all the way to the input *end*. Not only does this save huge amounts of researcher time, counterintuitively it also results in better performance when trained at suitable scale (with more data and parameters) [227, 542, 280]. This is because "credit" for improvements can be jointly assigned across layers, and so better representations are learned [56]. This loss of modularity is, however, yet another obstacle to the interpretability of DL systems' decisions, contravening as it does a key principle of good engineering [29, 309].

## 2.2 The Large Scale Era (2016–Present)

We date this second era in deep learning to 2016, following the quantitative analysis in [504] and the wild post-2017 success of massively scaled Transformer neural networks [570] and the AlphaGo-MuZero lineage of deep reinforcement learning (RL) systems [516, 499]. Figure 3 lays out some key factors that led to an exemplar large scale system, GPT-2 [460], and to emergently multitask systems with passable understanding of language and a broad range of abilities [582]. (Note that it is not possible to produce an equivalent diagram for more recent systems: AI development has "gone dark" in the intervening three years, with roughly complete papers replaced by technical notes which omit even basic information like parameter count [428].) Figure 4 presents heavily idealized differences between neural networks in each era.

In retrospect, we can view the Liftoff era as a search for a scalable "data sponge" architecture [271, 603] and a way to make use of unlabeled data, i.e. the vast majority of all data. The search succeeded: the dominant training scheme for cutting-edge models is now unsupervised (or self-supervised) pre-training followed by supervised fine-tuning, very often using the Transformer [68, 98, 104]. More importantly, this era established that parameter count is more crucial for performance than network depth and architecture [284, 61], which again foregrounds the role of hardware and engineering (in this case, breakthroughs in model parallelism which permit the inclusion of vastly more parameters).

*The Transformer and the pending unification of ML* In 2012, ML was divided into research communities which each worked on one type of input, generally using different methods and holding separate conferences. Each input modality involved different tradeoffs, different domain knowledge for hardcoding inductive bias, and different engineering challenges. Each of these communities focused on computer vision, natural language processing, speech processing, reinforcement learning, or probabilistic methods, and so on. Deep learning became dominant in computer vision first [310, 233], followed by word embeddings [385], reinforcement learning [392], machine translation [37], and speech processing [227]. After 2017, the worth of massively scaled neural networks was realized. Here, "scaled" refers to nets with greatly increased parameter count and the associated increases in training data size and training compute. The architecture most often scaled is the Transformer [570], but other architectures can reach performance equivalent to the Transformer's when equivalent dataset sizes and parameter counts are used [61]. Massively scaled nets enabled unprecedented progress on the input modalities mentioned above—and, crucially, on multimodal learning (the use and integration of information from several "senses" or data types at once, as in image captioning or text-to-speech audio generation) [269]. It is vital to tie these other domains to language because symbols index conceptual space and thus allow searches to be performed. Suitably scaled Transformers now achieve state-of-the-art performance in several domains: in order of application, machine translation [570], language modelling [146], computer vision [156], speech [96, 327], and multimodal combinations of the above [401, 601, 15, 472]. At the logical extreme of this trend are single pretrained models serving as "foundation models" for a huge array of downstream tasks and modalities [68].

*RL comes of age* Another impressive lineage of systems are trained using deep RL, with unprecedented performance and generality (including long-awaited success with model-based RL and real-world deployment) [392, 516, 499, 219, 39, 451, 171, 138, 287].

*Outlook* New benchmarks and challenging examples are now often beaten within months [435]. However, this trend relies on exponential increases in inputs [504] and is thus unlikely to continue for long without greater algorithmic progress [167]. A weakness of the current paradigm is the use of best-case performance as the dominant publication criterion, sometimes to the total exclusion of efficiency, robustness, and theoretical motivation [297]. Generally, no adjustment is made for the amount of optimization performed across methods [518], which means that impractical methods can be presented as state-of-the-art simply because one research group spent more. Similarly, it is normal not to report your method's sensitivity to random seeds [455].

Further, the practice of iterating on previous hyperparameters and selecting methods based on previous performance on the same task is common, but represents a subtle kind of data leakage (i.e., a violation of the assumptions of empirical risk minimization). Combined with overparameterization, this can make conventional DL unfit for tasks where overfitting or data shortages are a live concern [198, 351]. (For work focusing on improving the robustness of ML outputs, see Section 3.2.)

## 3 The Hard Problems

Here we present each of the AI2050 Hard Problems. For each, we present the results of our literature search, discuss what makes the problem hard, and examine prospects for future work in the area.

3.1 Hard Problem #1: By 2050, we will have solved the **scientific and technological limitations** in current AI that are critical to enabling further breakthrough progress and powerful AI capable of realizing beneficial and exciting possibilities.

Between 2010 and 2020, artificial neural networks (ANNs) re-emerged as the dominant tool in AI research, under the branding "deep learning" (DL) [500]. During that period, the publication rate of papers on ANNs increased thirty-fold [604]; at the same time, systems of unprecedented capability were trained through a *ten order-of-magnitude* increase in computational input (Figure 2). Compared with alternative approaches, the triumph of DL is that it productively enlists vast computational and data resources, and continues to meaningfully increase in capability as it scales up. The result is that today's best AI systems are billion- to trillion-parameter neural networks trained on terabytes of high-quality data using petaflop-days of compute [504, 21]. (See Supplement 2 for a more detailed explanation of the DL boom.) Despite remarkable successes in ML perception and the emergent capabilities of large language models (LLMs), these systems still lack reliability and persistence, and remain inscrutable owing to their high dimensionality and distributed representation, the black-box optimization processes used to train them, and the difficulty of reverse-engineering the circuits and algorithms they learn [424, 405].

To realize the potential of AI, the field will need to develop approaches that can be applied across datasets, domains, input modalities (e.g. pixels, text, waveforms, and control signals), and forms of cognition (e.g. raw sensory processing, language understanding [87], symbolic reasoning, physical movement, content creation, and the ability to learn and perform a wide range of tasks) [9, 371, 568, 582].

To demonstrate progress in a particular capability, machine learning researchers present an implementation with unprecedented results on some respected performance benchmark. However, when it comes to recent LLMs (and other "foundation models" [68]), these demonstrations tend to *lower-bound* the system's capabilities, because emergent capabilities that were not part of the training objective are increasingly being discovered as we increase parameter counts and improve our "prompting" (i.e., sampling) strategies [582, 74]. Recent examples of emergent skills include "showing your work" (step-by-step informal logical reasoning) [420, 582], high-quality text summarization [429], competency in basic physics, the ability to identify irony, and the ability to identify analogies [581]. (It is sometimes argued that this emergence is a mere artefact of the particular evaluation metrics used [494], but we note that this does not apply to bilingual evaluation understudy or to high-level "in-context learning" capability (runtime improvement on a huge range of tasks, given only examples and no further training steps) [79, 357, 582, 595, 225]. Barak further notes that difficult tasks involve the conjunction of many small tasks—and attaining all clauses of such conjunctions is binary, so capabilities on complex tasks can be legitimately emergent [42].) The result is a *capability overhang*, in which we do not know the limits of today's large models because it is infeasible to test them exhaustively [467, 427].

Following publication, it can take researchers years before they can explain how their work contributes to improved system performance (if their contribution is ever explained). The most commonly used explanations are ablation (the removal of elements of the system while observing the degree of performance degradation [381]) and retrospective theoretical analysis (in which we prove hopefully-relevant theorems about the method's time complexity or its equivalence to an ideal inference framework) [191, 247].

*Is deep learning enough?*   How much can DL continue to improve through increased computation [552], more and better data, engineering tricks, and patching in other software tools [69, 383]? Do we need a new paradigm? In their recent study of AI experts, Cremer identified five key disagreements about the limits of ANNs [119]:

**Abstraction:** "Do ANNs form abstract representations effectively?"
**Generalization:** "Should ANNs' ability to generalize inspire optimism about deep learning?"
**Causal models:** "Is it necessary, possible and feasible to construct compressed, causal, explanatory models of the environment... using deep learning?"

**Emergent planning:** "Will sufficiently complex environments be sufficient... to develop... long-term reasoning and planning?"

**Intervention** : "Will DL support and require learning by intervening in a real environment?"

To this we might add:

**Sample efficiency:** Will DL be able to learn from just a few examples, similar to the human brain? [311].

These are wide-open scientific questions. While some researchers predict a negative answer to one or more of these six capability questions [169], we note that, to date, improvements from large-scale DL have not stopped—and the rate at which new benchmarks are solved may also be increasing [535].

One tool for predicting the future of machine learning involves black-box "scaling laws" [284, 246]. These empirical curves are fit by varying the dataset size and number of parameters used to train a large model for a given compute budget, and they allow us to predict loss over different levels of input (dataset size, model size, and amount of training computation). Remarkably, scaling improvements have so far held over eight orders of magnitude of training computation [284]. The most recent such law [246] implies that a relative lack of training data is presently the limiting factor, rather than the engineering challenges of increasingly vast models or scarce computational resources [573, 358]. It thus seems likely that any future improvements in this paradigm will result from better algorithms, more data, and better data quality, rather than the brute-force scaling of parameter counts.

*Are AI capabilities a virtuous cycle?*  A much-discussed milestone in AI research is the point at which systems begin to contribute to the design of their successor systems—or, more generally, when AI research improves the efficiency of AI research. Such a recursive process would produce an exponential acceleration in capabilities [211, 70, 223]. The emergence of such a virtuous cycle would require success in one or more of these specific research components:

**ML optimizing ML inputs**  To date, ML systems have optimized several key inputs to ML R&D by reducing datacenter energy costs [171], reducing the cost of acquiring training data [24], and improving semi-conductor designs, including for AI chips [486, 312, 387]. The image data used to train deep learning systems has also been expanded using deep learning systems [511], possibly addressing concerns around long-term data quantity and quality [573].

**ML aiding ML researchers**  Similarly, code suggested by ML models has improved programmers' efficiency, with 3% of new Google code now being auto-suggested [545], and potentially an even larger share of GitHub repositories benefiting from this [152]. It seems likely that this already feeds into the increasing efficiency of ML research (or at least ML deployment).

**ML replacing ML research**  ML systems taking over tasks which were previously performed by hand is some-times branded "AutoML" [263]. This includes automatic data cleaning and feature engineering, automatic and symbolic differentiation [321], meta-learning of network components (like activation functions [12] and optimizers [379]), and neural architecture searches [606, 165, 345] in which a neural network designs better neural networks. In addition, steps have been taken towards automating fundamental discoveries, as with the discovery of a nominally better algorithm for matrix multiplication [175] (though this specific algorithm has a highly limited domain).

**Direct self-improvement (i.e. models which improve themselves)**  The most successful example is "self-play" (in which a system is pitted against a copy of itself, thus producing infinite training data and a smoothly increasing need for greater capabilities) [515, 224]. Recent work with unclear practical effect includes "Algorithm Distillation" (which automatically replaces RL algorithms with notably more sample-efficient versions [327]); and, most strikingly, language models which fine-tune themselves *on their own output*, producing reported improvements in programming and reasoning ability [224, 257]. A notable example is the Self-Taught Optimizer, which chains language model calls to improve a target program's programming ability [600].

The practical significance of some of these components has been contested [78, 196, 2], but they stand as early proofs of concept, while other innovations (like data augmentation and code completion) are already having a measurable effect on the productivity of researchers and engineers [545, 152].

*2024 update*  Of the Hard Problems, HP#1 moves so quickly that *some* update on our January 2023 view is mandatory. Besides a vast outpouring of consumer products (surpassing $1bn in revenue for the first time [162]), open-source initatives [393], and the landmark multimodal GPT-4 and Gemini releases [428, 197], in 2023 some progress was made on the scientific questions covered in this HP section. In our view, the "stochastic parrot hypothesis" (that LLMs only learn superficial correlations and do not have stable representations of the world, or any "understanding") is disconfirmed [344, 85]. However, we also have a clearer picture of the *degree* to which LLM outputs reflect "only" compression or memorization of the model's training data [141]. Recall that the training data used to create frontier LLMs are usually not public for competitive reasons [45]. As a result, it is difficult for external researchers to tell whether a given successful LLM output is more a result of reasoning than of recall. Let the *%-memorisation hypothesis* be the claim that some proportion of apparent LLM reasoning is in fact recall of hidden training data. The strong stochastic parrot hypothesis claims 100%-memorization, but most tasks appear to instead be 20-60% memorization (in the sense that this is the degree to which performance degrades on problems constructed to be novel, that the model has not been trained on) [591]. Finally, we should note that compression of training data does not preclude intelligence [141]. We also have a better understanding of LLMs' key "in-context learning" (emergent, few-shot) ability [582, 225]

*Summary*  An entire industry and a major academic field are focused on solving HP#1 (bound up with its cousin HP#4). Being upstream of applications, capabilities are the root source of *all* the value and risk associated with AI. Researcher incentives already lie in the direction of *best-case* performance, and the marginal attempt to accelerate capabilities may have less impact than work on the other Hard Problems, which is less well-funded and well-staffed. Worse yet, as many researchers have noted, increasing capabilities decrease the time available for society to acclimate to current AI systems and prepare for more advanced systems [513, 382, 380, 559, 364, 557]. These preparations include improving system robustness and human oversight (HP#2), preventing harmful emergent goals (HP#3), creating stabilizers for international relations (HP#6,7), and putting in place social safety nets (HP#5, 9).

## 3.2 Hard Problem #2: By 2050, we will have solved AI's continually evolving **safety and security, robustness, performance and output challenges and other shortcomings** that may cause harm or erode the public trust of AI systems, especially in safety-critical applications and uses where social stakes and risk are high.

While HP#1 concerns mean or best-case performance, HP#2 concerns worst-case performance: how can we ensure that AI systems will perform safely, and how can we prove this? ML systems have been implemented in high-stakes, safety-critical domains such as driving [182], medicine [113], and warfare [298]. Many more systems have been developed but have remained undeployed or been rolled back as a result of regulatory and safety reasons [471]. Clearly, unsafe systems can result in loss of life, economic damage, and social unrest [407, 10]. Most concerningly, AI systems may be susceptible to so-called "normal accidents" [63], creating cascading errors that are difficult to prevent merely by maintaining a nominal "human in the loop" [122].

Most advanced ML models perform far below the reliability level customary in engineering fields [359]—and because we do not fully understand how cutting-edge systems achieve their results, we cannot yet detect and prevent dangerous modes of operation [285].

Key approaches for solving HP#2 include *systemic safety*, *monitoring*, *robustness* and *alignment*. (See Figure 5 for a conceptual illustration.) We discuss the first three here, and alignment in HP#3. In addition, the subfield of *formal verification* of neural networks has begun to make progress on smaller models. These methods produce a proof that the system meets a specification [529]. If successfully scaled (and if applied to safety properties at a much higher level of abstraction than the usual input-output functions), these methods could help catch unsafe systems before deployment.

*Systemic safety*   Traditional software security methods, including formal methods, could help ensure that AI systems are run in secure environments and constructed using trustworthy software. However, such approaches are incomplete and unlikely to be perfected even by 2050. This aligns with a key insight of safety-critical engineering: safety is a social problem as much as a technical one, because systems are always deployed in a social context which determines their actual performance. Safety thus requires careful design and vigilance on the part of developers and users [341].

Some researchers argue that HP#2 can be addressed by simplifying ML systems so that they are more amenable to verification and validation. This assertion relies on results showing that, in some domains, simple models can match or approximate the performance of more complex systems like neural networks [488, 419, 502].

*Monitoring*   Public trust in AI systems requires that we be able to observe, interpret, and ultimately understand AI outputs. Approaches for understanding these incredibly complex systems include *mechanistic interpretation* (determining exactly what function a subset of a system is executing); *concept-based interpretation* (e.g. "this is the collection of neurons which correspond to 'forking' when the system analyzes a chess game", etc.); and *feature-based interpretation* (e.g. "this square is critical to the system's evaluation of the chessboard", or "these pixels are critical to the system's attempt to determine the number of cats" [377].

An ideal interpretability tool would directly report the internal representation used by the system in a relatively faithful low-dimensional form [102, 423, 577], but building a system that is fully interpretable (as advocated by e.g. Räuker et al.[469]) likely requires building interpretability into the training process (by choosing models legible to humans) or paring down more complex neural networks (as in model distillation or mechanistic interpretability) [16, 577].

*Robustness*   Building a system which performs properly or fails gracefully in all conditions requires us to recognize and avoid areas where our system will fail. To build such a robust system we might drastically perturb training dataset with various forms of noise [235]; ensure adequate diversity in the training data; and "red-team" the AI system with adversarial examples generated by humans or by other AIs [450].

Correct model calibration helps ensure safety. However, while modern algorithms have greatly improved average performance, they are consistently overconfident compared with ground truth and with simpler models—even on inputs drawn from the same source as their training data [216, 279]. When inputs are no longer sufficiently similar to the training data, the system is said to be "out-of-distribution" (OOD). As systems go further OOD, their performance degrades. All offline AI systems are somewhat OOD, simply because distribution shift is omnipresent in real-world contexts: systems are trained on past data and deployed in the present. It is increasingly possible to automatically detect when new data is OOD, allowing systems to acknowledge this fact and so fail safely [359].

*Summary*   Any solution to HP#2 will require systematic testing. Realizing the potential of AI will thus depend on our constructing and using strict and credible tests of real-world ML behavior.

To date, most testing of AI systems has used benchmarks which have been criticized as artificial and easily gamed [400, 194, 549, 466]. As the computer scientist Boaz Barak puts it: "The history of artificial intelligence is one of underestimating future achievements on specific benchmarks, but... overestimating the broader implications of those benchmarks." [43]. More recently, some researchers have focused on system performance "in the wild" (e.g., driving through city streets) and through "adversarial testing" by human and ML critics [194, 292]. Meanwhile, third-party AI testing is emerging as an attempt to check that AI systems do not engage in dangerous actions (like deception or self-replication) [17] or discrimination against legally protected groups [288].

### 3.3 Hard Problem #3: By 2050, we will have solved the challenge of **safety and control, alignment, and compatibility** with increasingly powerful and capable AI systems and eventually those of artificial general intelligences (AGI).

How do we ensure AI acts according to our values? Equivalently, how do we prevent poorly-understood AI systems from advancing goals we do not endorse? Whereas HP#2 concerns the prevention of harm caused by incompetent systems, HP#3 seeks to align *competent* AIs with humans, through methods which ensure their behavior is compatible with the user's intentions.

Of course, HP#3 is complicated by the lack of agreement among humans about values. Different cultures have different approaches to risk tolerance and different definitions of harm, and so different regulatory environments have different legal standards for how automated systems should perform when they, for example, encounter OOD inputs [359]. One goal which is robust to normative variety is *intent alignment*: producing a system which is at least trying to do what the user intends it to do [188, 340]. Figure 6 shows one decomposition of the alignment problem.

Figure 7 depicts the "greater alignment problem" of economic and strategic barriers to the successful use of alignment methods once we have established them. This can be viewed as a central problem of AI governance: see HP#9.

**For an updated (2024) view of approaches to the alignment problem, see our informal review [332].**

*Problem: Specification gaming*    When the US adopted standardized testing to determine the quality of public schools, educators predictably responded by "teaching to the test": under-emphasizing any part of the curriculum that was not included in the exam [273]. Likewise, when the UK National Health Service adopted a stringent target that all emergency patients would be seen within four hours of entering hospital, the result was columns of ambulances idling outside hospitals for hours to prevent the target timer starting [62]. These cases can be described as cheating, because the metric is being optimized at the expense of the underlying purpose.

Similarly, AI systems game specifications [305]. For example, in 2017 an OpenAI robot trained to grasp a ball via human feedback from a fixed viewpoint learned that it was easier to pretend to grasp the ball by placing its hand between the camera and the target object, as this was easier to learn than actually grasping the ball [103]. Researchers may respond to specification gaming by adding detail to the specification of the desired task (in the form of a more complicated objective function, or more labeled instances, demonstrations, or advice) [255]. However, it is difficult to specify *all* the relevant desiderata and conditions, even when using a learned implicit model of human preferences [22, 491]. Optimized against an incomplete specification, a system will find loopholes and may instead perform a spuriously correlated task [549]. Such behavior can be called *specification gaming* or *reward hacking* [440, 519, 112], and has now been observed in many AI systems [305, 549].

A natural response is to simply add variables to the specification when it is revealed that they are needed—an interminable process, owing to the sheer number and context-sensitivity of human preferences and the difficulty of specifying them formally [491, 22]. As Russell argues, "It is... perhaps impossible, for mere humans to anticipate and rule out in advance all the disastrous ways the machine could choose to achieve a specified objective." [490] So one part of a solution to HP#3 is to detect flawed specifications and cheating, and to find ways to specify robust objectives [260].

*Problem: Emergent goals*    As well as optimizing a subtly wrong goal, systems can develop harmful instrumental goals in the service of a given goal—without these emergent goals being specified in any way [434, 218, 339, 17]. For instance, a theorem in reinforcement learning suggests that optimal and near-optimal policies will seek power over their environment under fairly general conditions [560]. This power-seeking behavior is plausibly the worst of these emergent goals [92], and may be an attractor state for highly capable systems, since most goals can be furthered through gaining resources, self-preservation, preventing goal modification, and blocking adversaries [426, 449]. Presently, power-seeking is not common, because most systems are unable to plan and understand how actions affect their power in the long term [414].

A further risk is presented by systems that deceive us about their alignment. Current systems can learn to deceive: for instance, the Cicero system plays a version of the game Diplomacy at human level, employing persuasion, deception and betrayal despite being trained for honesty [40]. Perhaps the greatest risk involves *deceptive alignment* [91], in which a system learns to detect human monitoring and hides its undesirable properties—simply because any *display* of these properties is penalized by the feedback process, while that same feedback is usually imperfect. (Consider the problem of verifying a translation into a language you do not speak, or of checking a mathematical proof that is thousands of pages long.) [92, 259]. Rudimentary examples of deceptive alignment have been observed in current systems [322, 333].

*Approaches* For a review of current research agendas in AI alignment and control, see our informal work [332].

Currently, the dominant approach to HP#3 involves iterated human feedback: humans reward systems that output good behavior, and penalize unwanted behavior by making significant modifications to the system. (In practice this process requires so much feedback that it must be automated by using a proxy model of human preferences [103, 436, 204, 38, 338].) Again, this feedback method selects for systems that *appear safe* during training, and thus leaves open the possibility of unsafe systems which merely appear safe. In particular, if a system exceeds some threshold of planning ability, seemingly innocuous feedback might simultaneously penalize misbehavior and incentivize deception [5]. The system could thus pass behavior tests, but still deviate whenever it could do so without being detected. A naive response is to simply severely penalize even mild forms of such behavior—but this only delays the problem, since such penalties greatly increase the selection pressure towards opaque and patient deception [70, 115]. The response also fails to apply to tasks where human scoring of results is imperfect or impractical (as with any long-range task with slow feedback loops or messy causal inference).

The leader of the alignment team at OpenAI, Jan Leike, has proposed that parts of the alignment problem be delegated to ML systems, particularly the generation of ideas and the design of scalable systems [335, 334]. Similarly, Christiano has proposed an AI interpreter for AI systems [102]: an advanced, adversarial form of dimensionality reduction which discovers the target system's high-level representations. If achieved, this Eliciting Latent Knowledge (ELK) system could help detect and train out misalignment. Perez et al. use language models to judge the output of other language models, speeding up the process by more than an order of magnitude [449].

*Summary: AI assurance, AI ethics, and AI alignment*   Researchers working on the different risks posed by AI overlap in personnel, methods, and goals [267, 121, 238]. Alignment research thus overlaps with assurance (HP#2) and responsible AI (HP#7): each involves mitigating risks to hard-to-model, high-level features of the human environment. In particular, the alignment problem can be viewed as a kind of robustness problem: we need to ensure that systems robustly generalize about the goals we are training them to fulfill [323, 238].

Despite this, there is some debate over the proper horizon with regard to harms: should we focus on current systems and current risks, or look ahead to future risks? [331] This is a false dichotomy: AI risk is better viewed as a continuum, both ends of which are worthy of concern [458, 94, 538, 48, 189]. Consciousness of one risk does not appear to trade off against consciousness of other risks [214]. Aligning current systems against current and foreseeable harms plausibly provides us with the best feedback we will get for aligning stronger systems [114, 31, 540].

3.4 Hard Problem #4: By 2050, there will have been game-changing **contributions by AI to humanity's greatest challenges and opportunities**, including in health and life sciences, climate, foundational science (including the social sciences) and mathematics.

Researchers across disciplines and industries are applying AI to their hardest data analysis and discovery problems. Many of these, like improvements in energy and healthcare, are crucial for the future of humanity. These applications include fundamental discoveries, building AI into beneficial applications, and using AI to accelerate the pace of R&D and technology transfer. Already, automatic differentiation tools have reduced mathematical drudgery in science, technology, engineering and mathematics (STEM) fields, including ML [153, 50]. Some applications are overlooked because they succeed to the point of becoming routine background: consider "prediction machines" [13] like spam filters [531], recommender systems [457], the turn to ML pricing across industries [65, 584], or the ubiquity of serviceable machine translation (including real-time translation from images and speech) [445]. This section briefly explores progress in select areas; see [464, 133, 49, 474, 26, 136, 325] for more.

*Scientific discovery*   ML has led to remarkable scientific breakthroughs in recent years, including the AlphaFold system increasing the number of predicted protein structures by two orders of magnitude [230]. One worry is that these systems give us answers but do not increase our understanding; but Krenn et al. sketch ways in which systems could aid us in this too [308]. One example of ML assisting discovery is models substituting for exact simulations, as in Figure 8: exact simulation of quantum systems is incredibly computationally expensive, but can generate perfectly labeled training data to train machine learning proxies, which can then support much larger searches and larger quantum systems [256, 307, 530].

*Energy & climate*   AI applications in clean energy have garnered significant attention and investment. Rolnick et al. list many such applications [483]: optimizing electricity grids, enhancing transportation efficiency, refining energy usage in heating and construction, streamlining industrial supply chains, carbon removal techniques, and improving climate models. AI has also contributed to fusion research [262, 138]. The interdisciplinary organization Climate Change AI was founded to focus on this strand of applications [111].

*Chemical and materials research*   A fundamental challenge in the design of new catalysts, batteries and advanced materials is the modeling of complex quantum wave equations that might involve hundreds of atoms and thousands of electrons. Recent ML systems can perform elements of quantum simulation [256] previously thought to be solely the domain of much-anticipated quantum computers [597, 408]. Meanwhile, quantum computers may come to depend upon machine learning for error correction [306, 569].

*Software development and algorithm design*   Language models optimized to output source code [224, 97] are reportedly now writing 3% of all new code at Google [545], and perhaps an even larger fraction of Github repositories [152]. Currently, humans remain in the loop, reviewing code completions and fixing the (common) bugs. Language models continue to show remarkable improvements in programming ability, achieved for instance by bootstrapping with a compiler or an interpreter as a training signal [224]. The Codex model (a heavily fine-tuned GPT-3) generated bug-free code 29% of the time on a small benchmark of basic problems, while a baseline GPT-3 passed 0% of these problems [97]; the AlphaCode system solves difficult competitive programming tasks at median human level [599]. More interesting still is early work on algorithm design, for instance the use of deep RL to discover a novel efficient algorithm for matrix multiplication (though over a very limited domain) [174].

*Healthcare*   High variance in ML performance remains a serious obstacle for high-stakes applications like healthcare. In 2016, the deep learning pioneer Geoffrey Hinton famously claimed that "we should stop training radiologists. It's just completely obvious that within five years, deep learning is going to do better than radiologists" [20]. The evidence suggests that this prediction was overstated rather than outright false.

As of 2022, around 40% of European radiologists are using AI tools [51]. This is far short of the full automation implied by Hinton, and does not strictly imply that the tools are clinically useful—but this is still remarkable penetration in only six years. We found little data on the promised cost savings of ML healthcare; one study found a 10-19% cost reduction for one procedure [395]. Similarly, the first drugs suggested by AI systems predicting clinical relevance from chemical structure are now entering human trials; there is a good chance this trend will revolutionize drug discovery [95, 282, 352].

Despite these encouraging signs, uptake remains slow. How can this be explained? One concerning review shows that bad methodology is rampant in healthcare ML: only 12.5% of diagnostic and prognostic studies included a test set, and only 10% performed any calibration analysis [23]. Of 232 Covid diagnosis models selected for relative *high* quality by Wynants et al [592], *all* had a "high or unclear" risk of bias, and only two passed basic performance tests. Roberts et al [482] similarly find that "none of the [62 imaging] models identified are of potential clinical use due to methodological flaws and/or underlying biases". These problems are found in other domains [347, 439], and likely explain a large part of the deployment gap.

*Summary*    The pattern is that fields with large amounts of data, strong theory, and relatively stationary distributions benefit most from ML, just as they benefit most from mathematization and statistical modelling. Similarly, the AI products which are currently generating revenue are mostly in areas with imprecise specifications and a low cost of error—consider art, copywriting, customer service, code completion with a human in the loop, and conversation bots [554, 432, 593, 545]. The low reliability of most ML methods is a general obstacle to deployment (see HP#2).

Further, in many of the above cases we have approximately no mechanistic understanding of *how* our successful systems succeed [577]. This is an uncomfortable situation for scientists, or anyone concerned with insight, control, and accountability. Looking ahead, we need applications that actually improve society. This will require progress on assurance (3.2), alignment (3.3), and responsible AI (3.7). (See Section 3.7 for possible ML applications for advancing the 'social good': e.g. AI countermeasures against poverty, pollution, and human trafficking [356].)

## 3.5    Hard Problem #5: By 2050, we will have met the **economic challenges and opportunities** resulting from AI and its related technologies.

Are the robots and generative AIs coming for our jobs—or will AI create new economic opportunities for humans that we cannot yet imagine? History is an imperfect guide here, given the apparently fundamental difference between AI's impacts in the past and its likely impact over the next 25 years.

At the onset of the Great Depression, John Maynard Keynes famously claimed that, in the long run, humanity's major challenge would be deciding what to do with our leisure time—since by 2000 we would work no more than 15 hours a week; "Thus for the first time since his creation man will be faced with his real, his permanent problem—how to use his freedom from pressing economic cares, how to occupy the leisure, which science and compound interest will have won for him, to live wisely and agreeably and well." [290]

Keynes did not foresee AI completing the centuries-long project of labor automation: he simply had faith in the power of compound interest and steadily increasing levels of productivity.[5]

The dissenting view was articulated by a US national commission in 1966: "The basic fact is that technology eliminates jobs, not work" [73]. This belief is often proclaimed by those in favor of automation, as opposed to those who stand to lose their current jobs. It also assumes a balance between the rate of job creation and job destruction—a balance that AI may disrupt, as Brynjolfsson and McAfee forecast:

---

[5]Keynes also assumed "no important wars and no important increases in population" would detract from the power of compound interest, and did not account for steadily increasing expectations on the part of consumers. Thus, today we find ourselves in a rich but labor-intensive world: since Keynes was writing, working hours have fallen about 25% rather than the 80% he predicted [202].

> Rapid and accelerating digitization is likely to bring economic rather than environmental disruption, stemming from the fact that as computers get more powerful, companies have less need for some kinds of workers. Technological progress is going to leave behind some people, perhaps even a lot of people, as it races ahead. As we'll demonstrate, there's never been a better time to be a worker with special skills or the right education, because these people can use technology to create and capture value. However, there's never been a worse time to be a worker with only 'ordinary' skills and abilities to offer, because computers, robots, and other digital technologies are acquiring these skills and abilities at an extraordinary rate. [83]

One famous estimate of automation risk by industry [186] illustrates the difficulty of forecasting such risks: creative professions like "visual artist" were only included in the paper's appendix, and were given a mere 4% risk of automation—but today, after a decade of generative AI development, commercial art has turned out to be at much higher risk. OpenAI researchers project that 80% of the US workforce might find 10% of their work tasks "affected" by OpenAI's GPT-4 system, and 19% of the workforce may see "at least 50% of their tasks impacted" [164].

We break this "hard problem" into three: (1) predicting the impact of AI on economic growth; (2) predicting its impact on the labor market and inequality; (3) possible policies to maximize beneficial outcomes.

*Predicting AI's impact on economic growth*    Despite continuing technological progress the key measure of economic growth, 'total factor productivity' growth, has been slowly declining in the developed world for 50 years, down to 1.0% per year in 2016, compared with 1.5% in the 1970s [118]. This reduced acceleration in developed economies is known as the *productivity paradox* or *Solow paradox*. In 1987 Solow himself wrote that "You can see the computer age everywhere but in the productivity statistics" [526]. Three decades later, Brynjolfsson, Rock, and Syverson noted that the paradox still held, attributing it to the lagging diffusion of AI technology. Other technologies (e.g. electrification, semiconductors) averaged 25 years of slow growth before having profound impacts on productivity, and we might expect AI to follow a similar pattern [84].

There is no consensus about how AI effects should be incorporated into growth theory. One paper finds 25 distinct modeling choices in the literature [556]. Most models formalize AI as a capital-augmenting factor [6]—but some models include it as a labor-augmenting factor [60] or a replacement for highly-skilled labor that simultaneously generates innovation [11]. Starting afresh, Trammell and Korinek [556] examine four ways in which AI could transform economic growth:

(1) "a decrease to the growth rate, even perhaps rendering it negative";
(2) "a permanent increase to the growth rate, as the Industrial Revolution increased the global growth rate from near zero to something over two percent per year";
(3) "a continuous acceleration in growth, with the growth rate growing unboundedly as time tends to infinity";
(4) "an acceleration in the growth rate rapid enough to produce infinite output in finite time."

The last scenario is physically impossible, of course, but the relevant question is whether the AI transition is better modeled using this drastic functional form rather than *brief* periods. According to Trammell and Korinek, there are no compelling theoretical reasons to dismiss such scenarios [556]; in principle, advances in robotics and AI allow for super-exponential growth [493].

*How will AI affect labor?*    Following long-held concerns about automation, the most common AI-related topic among policymakers and the public regards AI's threat to jobs. The current evidence is flawed: most empirical studies of AI-induced unemployment use a definition of AI (as mundane automation) that covers only a subset of the potential effects of AI on the labor market. One widely cited paper claims that "up to 35% of all workers in the United Kingdom, and 47% in the United States, are at risk of being displaced by technology over the next 20 years" [186]. This could result in substantial employment displacement, exacerbating income and wealth inequalities [303] in the absence of corresponding social programs. The first empirical studies on the effects of LLMs on industries have begun to emerge, showing notable increases in productivity among e.g. median customer support workers [82]—and also associating wage decreases with automation in general [7].

Acemoglu and Restrepo note four factors that may instead contribute to a *positive* effect on employment:

(1) *The productivity effect*, whereby cost savings from automation reduce prices and so increase consumer demand (in the sector experiencing automation or in other sectors). This could increase the demand for labor to perform the remaining non-automated tasks.
(2) *The capital accumulation effect*, whereby automation increases the capital intensity of production, triggering accumulation of capital, which also raises the demand for labor (in tasks where AI is complementary to human labor);
(3) *The deepening of automation*, whereby technological improvements increase the productivity of existing machines with no additional displacement of labor, boosting the productivity effect and further increasing the demand for labor; and
(4) *New labor-intensive tasks* made economically feasible by AI assistance, which could increase the labor share of income (an effect which could potentially persist in the longer term), which would also counteract the impact of automation.[8]

*Possible interventions*    Realization of the benefits of AI requires the mitigation of any negative economic impacts. It is increasingly plausible that this mitigation will itself involve the use of AI [301].

Economists were early adopters of machine learning methods, with some notable improvements to empirical methods [64]. One simple example is the Bank of Italy's use of natural language processing to track inflation expectations by analyzing millions of Twitter feeds [25]; the results are reportedly more accurate than other sources.

As with other areas, reliability is a serious obstacle. For example, the Polish Ministry of Labor used AI to automate its unemployment benefit process, assigning individuals to categories (e.g. job placements, vocational training, apprenticeships, allowances); however, the system had to be dismantled following widespread reports of inaccuracy [317]. Similarly, well-cited work using Twitter to predict the stock market [67] failed to pan out: "the hedge fund set up in late 2010 to implement the Twitter mood strategy, Derwent Capital Markets, failed and closed in early 2012" [319].

AI systems like the experimental "AI Economist," a multiagent RL system able to set taxation policy in a simulated economy [605], could help make policy decisions. This particular simulation has not been used in real policymaking, but successor methods could help vet decisions.

If it becomes necessary to mitigate the downsides of AI through policy, what will we do? Radical changes such as universal basic income could result in net welfare gains, but uncertainty remains worryingly high [229, 583]. Governments might contrive incentives for companies to hire people (and for workers to get hired) [302]. This might also help us retain the considerable noneconomic benefits of meaningful work, such as dignity and social cohesion.

It is important to emphasize the unpredictability of AI capabilities, and therefore of the labor market shocks associated with them [582]. None of the classic studies of AI automation foresaw 2022's nascent automation of commercial visual art [485]—even though the breakthrough followed one study (Nedelkoska and Quintini) by less than four years [410]. We can say with some confidence that capabilities will increase, that these increases are likely to be sudden and unexpected, and that few social institutions will be prepared for their effects.

### 3.6 Hard Problem #6: By 2050, we will have met the challenge of democratizing **access, participation, and agency in the development of AI** across countries, organizations, and segments of society, especially those not presently involved in the development of AI.

Our next problem is the fact that the current AI workforce does not evenly represent world demographics. Men from the US and China, working in the US, for US corporations, are disproportionately highly represented [402, 157, 170, 534]. Realizing the full promise of AI requires that people throughout the world and from all social strata are able to use AI and participate in its design and governance. Solving this problem requires addressing unequal access to AI both within countries and across countries.

#### 3.6.1 Within-country issues: domestic inequality

*Demographic diversity of researchers*    The AI research establishment inherits patterns of under-representation that are dominant in most technical fields. In North America, large parts of professional AI research require a Ph.D., yet less than 25% of Ph.D. computer scientists are women, and fewer than 2% are Black or African American [608]. This holds globally and outside the research community: LinkedIn data suggests that only 22% of AI professionals are women [161]. Since the vast majority of AI practitioners work for private companies, limited corporate statistics on gender and racial diversity hinder a full understanding of the situation [402], but those few statistics that exist are not encouraging: only 5% of Google and 7% of Microsoft employees are Black or African American, with potentially even lower representation at the more senior levels [212, 384].

A report by the European Institute for Gender Equality cites several reasons for gender discrimination in the specific field of AI, examining barriers to women's entry to the field through education (e.g. gender stereotypes and educational choices) and the workplace (e.g. sexual harassment, male-dominated teams, and lack of access to funding) [170].

We should not assume that improving the representation of under-represented groups in the field will necessarily improve the final product for those groups; some authors warn against "participation washing," in which nominal participation provides legitimacy to a project regardless of the actual outcome [521, 520].

*Privatization of AI*    Over the past two decades, there has been a net migration of AI researchers from academia to industry [278]: a study of North American publications found that the researchers behind 19% of *all* AI citations moved to industry between 2000 and 2018 [210]. Similarly, Ganguli et al. found that the share of 'large-scale' AI systems run for academic purposes fell from approximately 70% in 2000 to approximately 15% in 2020 [192]. This trend is likely to be self-reinforcing, as private incentives for non-cooperation and the keeping of trade secrets, combined with the associated centralization of decisions and benefits, will likely make AI research more expensive and less rewarding for those who remain in academia [278].

Researchers in deep learning and those with greater research impact are more likely to migrate to industry, raising concerns about the "privatization of AI knowledge" [278]. Specifically, if the most sophisticated AI approaches become proprietary and are used only within private research labs, then it will be impossible for universities to teach them, let alone contribute to leading research.

One reason for this shift is the intense capital requirements involved in staying at the frontier of research in the "large scale era" of deep learning (see Section 2). For example, the pretraining cost for Google's 2020 T5 model reached $10 million in compute costs alone [506]; we do not know how much more was spent on data annotation, software engineering, and other factors. If AI increases returns on capital, then it will tend to concentrate power in fewer hands [283, 35].

Even for those who remain in academia, the influence of private funding is significant. 58% of AI ethicists at four leading US universities have received financial support from major tech firms [4]. Some researchers claim that this helps tech companies disproportionately influence discourse, including decisions about which technologies get developed and adopted [222, 297].

The hybrid nonprofit/corporation OpenAI has released an API which allows those without giant compute clusters to use the flagship GPT-4 language model [430], and this currently serves as the *de facto* baseline in large parts of natural language processing research. However, the API provides no details on the training setup or specific model version exposed each time it is used (beyond crude top-level versions like 'davinci-001' vs 'davinci-002'), and these frequently change silently, rendering the original research papers non-replicable [336]. This *precludes* a scientific approach to our experiments, since we do not know the exact conditions of our baseline, cannot reproduce past results, and so cannot resolve disagreements about capabilities.

Against this trend, support for AI access comes from unexpected places. A 2020 report from the US National Security Commission on Artificial Intelligence (NSCAI) recommended the creation of public infrastructure to democratize AI research [496]. Picked up in the subsequent US defense budget, this led to recommendation of the creation of a National Artificial Intelligence Research Resource (NAIRR), a public cluster with the goal of "spurring innovation, increasing the diversity of talent, improving capacity, and advancing trustworthy AI." [441]

*Public participation*    Fixing AI workforce demographics helps promote AI access, but is not sufficient. Most people are not AI developers, and even research teams that are balanced by gender, race and other demographic factors create technologies that do not promote equal access and beneficial use. "Participatory technology assessments" provide a more structured approach to keeping developers aligned with the goal of equitable access as well as the impacts of AI on affected groups; such participatory frameworks have been used in fields such as 'climate change futures [248], and have even been trialed for highly technical projects [120, 446]. "Citizen Assemblies," in which non-experts are randomly selected and given sufficient background information to make substantive decisions about technical issues, have also been proposed as a way to broaden the governance of AI [181]. However, use of these frameworks is still in its infancy in AI.

*3.6.2    Between-country issues: global inequality*  There is an even greater divide between the countries currently leading in AI and those falling behind. While AI is widely considered a national priority, with almost 40% of countries having created an AI strategy [437], the implementation of these strategies depends on scarce resources, including trained STEM talent and computing power. These resources are predictably concentrated: 59% of leading AI researchers currently work in the US, and another 20% in China and Europe [372]. Figure 9 shows post-college migration among AI researchers who have published at one top conference, as of 2019.

A 2021 survey by the US National Science Foundation found that of the 1334 students graduating from a US institution with a Ph.D. in computer and information science, only 124 (9%) had plans to leave the country [289]; that same year, the Taulbee Survey found that 69% of computer science PhDs were awarded to graduates in the "Nonresident Alien" category [608].

*Structured access*    Open-sourcing ML tools and model weights is a powerful way to allow less wealthy users to catch up. However, this laudable aim sometimes conflicts with other aspects of responsible AI. Generative AI systems can assemble dossiers on individuals based on leaked personal information, and can produce limitless amounts of customized hate speech; so too can image generation systems produce synthetic revenge porn or instructions for the manufacture of weapons, among other harmful material [532]. Some organizations publishing these models have attempted to limit such uses [79, 477, 525], but other systems (such as the independent "GPT-4chan," which was trained on a notably bigoted internet corpus and then open-sourced [314]), intentionally have no such limitations.

Hosting generative systems and allowing access via a web browser or other API allows users without considerable computational resources to use cutting-edge models, and allows nontechnical people to access the models from any client device [525]. This use of APIs also allows generated material to be examined for dangerous or undesirable content (e.g. bomb-making instructions) and for such content to be blocked. This "structured access" model allows controls to be revised after problems come to light [508]. However, such access can also be terminated for less benign reasons, such as censorship or political sanctions.

3.7 Hard Problem #7: By 2050, we will have solved the challenges and complexities of **responsible research, deployment, and sociotechnical embedding of AI** into different societies and subcultures, accounting for different cultures, participants, stakes, risks, societal externalities, and market and other forces.

Many academics, community activists, artists, creative professionals and philosophers have raised ethical and social concerns regarding advanced AI systems.[6] These concerns include the unauthorized use of protected intellectual property as training data; poor working conditions for human annotators; the tendency of AI systems to exhibit racial or gender biases [402]; their ability to generate misinformation [143]; and their growing use for social control in authoritarian states [123]. Improvements in AI capabilities are likely to further exacerbate these concerns and create new ones [585, 366].

*3.7.1 Negative Impacts of AI Use* A major role of the current AI ethics movement is to draw attention to overlooked side-effects, costs, and harms of building and deploying AI systems, particularly as they befall existing marginalized groups:

- *Under-recognized work.* Without training data, ML cannot take place. Much of this data comes from paid clickwork (also called "platform work" [170] or "microwork" [558]), unpaid crowdsourcing, and unpaid user behavior capture. Clickworkers, mainly in the global south, perform repetitive data-labeling tasks for use in the training of ML models [558]. The market value of such annotations "is projected to reach $13.7 billion by 2030" [228] and the annotation industry is widely reported to have little concern for workers' rights. Besides welfare and rights, the invisibility of this contribution arguably contributes to a misunderstanding of AI capabilities.[7]

- *Environmental cost.* Large-scale DL systems can produce significant carbon emissions as a result of the computational demands of training runs and inference [539]. While AI has the potential to improve energy efficiency and progress on sustainable energy [116], estimating the trade-off between these two factors is complex and requires life-cycle estimations of carbon emissions, which have only recently been attempted [362]. Previous estimates focused on emissions from training systems [151, 41]; recent work has found that the majority of emissions may in fact be due to inference (the actual use of AI systems) [448]. Estimates are limited by the lack of transparency around the utilization of datacenters and hardware, and the difficulty of retroactively sourcing such data [447]—for instance, in at least one case external estimates of training emissions were off by two orders of magnitude [539, 447]. Additionally, the projection of current carbon emissions into the future is unreliable without an ability to model efficiency improvements in AI architectures, algorithms, accelerators, and datacenter usage, as exemplified by the 747-fold reduction in carbon emissions between the Transformer and Primer training setups [448]. Calls have been made for research papers to explicitly include measurements of carbon emissions [447], and to treat energy efficiency as a core component of competition benchmarks and conference awards [448, 391].

- *Discrimination, toxicity, and bias.* AI models and the tools that use them may exacerbate unequal access to employment and services. AI-generated content can promote inequality and harmful stereotypes. While proponents of AI systems argue that ML has the potential to remove racist biases in hiring or lending, critics claim that, in practice, these systems contain biases that require special attention from their human users [299]. For example, one system used to catch welfare fraud in Michigan had a 93% false discovery rate (almost all accusations of fraud were found to be mistakes, on review) [1]. The negative impacts of mistakes can be mitigated if negative decisions can be easily identified, appealed and rapidly reviewed; in practice such appeals are frequently impossible, such as when a classifier decides against showing a submitted resume to a hiring manager. Although there is broad consensus that preexisting bias in datasets is a significant source of undesirable behavior, there is heated disagreement about other sources of bias in machine learning, which could include sensors [342]; feature selection; and even biases rooted in gendered human language [374].

---

[6]This section omits the question of whether sufficiently advanced AI systems ought to be treated as moral actors. For full-length treatments of these questions, see [213, 215, 512].

[7]Note that this is only the most distinctive form of labor in AI; as a high-tech industry, AI also relies on the often hazardous work involved in procuring raw materials and manufacturing components [286].

- *Privacy.* Figure 10 shows a strong consensus in favor of respecting privacy. The qualified success of data protection laws such as the General Data Protection Regulation (GDPR) [199] and the California Consumer Privacy Act (CCPA) [88] has made privacy one of the areas of responsible AI subject to enforceable legislation [389]. Similarly, a growing body of research addresses privacy concerns [510, 396, 346]. Methods such as federated learning [300] and differential privacy [3] promise to allow ML training that incorporates tunable privacy protections with respect to training data extraction, but may not provide for protection of facts that are in the public domain but are scattered, such as those that may be included in dossiers created using large language models. OpenAI's GPT-3 was designed to be difficult to extract personal information from, including for example public figures' dates of birth. Even so, malicious uses of AI continue to encroach on privacy, as exemplified by China's "Sharp Eye" automated surveillance system [551] and automated cyberattacks on personal data [354]. A more drastic form of AI-enabled surveillance could be on the way in the form of nonsurgical decoding of thoughts [54]—a technique which is reportedly already used by some police forces [398].
- *Security* There is growing concern that AI-based systems can discover and exploit vulnerabilities in software or cyberinfrastructure [354].

*3.7.2 Principles for Responsible AI* Major institutions like governments and international organizations have responded to ethical problems with AI with a flurry of principles and guidelines: one review found 84 such documents, mostly published since 2016 [274]. Figure 10 summarizes the themes of these guidelines and identifies common and neglected themes.

Our analysis finds that few of these documents cover HP#3 or HP#10, express caution about dangerous capabilities or weaponization, present a technical research agenda to put principles into practice, or pay attention to future AI systems. Industry principles explicitly stress the benefits of AI capabilities. Google and DeepMind explicitly disavow working on AI-enabled weapons and surveillance systems, but most other documents omit the issue or stop short of prohibition. Principles written by industry entirely omit HP#5 (the future shock to the labor market), HP#9 (regulating AI systems), and environmental sustainability. Subjects underemphasized by governments include HP#8, specifically the impact on geopolitics, international cooperation, and approaches for avoiding racial conflict.

Statements of principle naturally lack detail about specific problematic technical features of AI systems—but they also lack specific mechanisms for ensuring that their principles are respected [220, 537, 598]. As Figure 10 shows, statements of principle are usually orphaned, with no clearly associated technical agenda or policy proposal to move them towards implementation and impact. Documents generated in industry and government are an exception. In common with [220, 390], we find that these documents systematically overlook certain topics, particularly future systems and the risks of increased AI capabilities.

We can make clear the limits of statements of principle by comparing their laudable aims to how AI is used in practice. For instance, the AI principles published by a Chinese state committee read "These norms aim... to promote fairness, justice, harmony, and security while avoiding such problems as bias, discrimination, and privacy." This is in stark contrast to the widespread use of AI in human rights violations in China [397, 578]. Ethics research that distracts from unethical actual uses of AI has been called "ethics-washing" [576, 598].

Funding allocations and research outputs also indicate which ethical constraints are active. Despite publishing twice as much work on AI as the US [604], Chinese research accounts for 1% of all research into social bias in AI; the US accounts for 48% [465]. Conversely, Chinese research constitutes 58% of the world's state-of-the-art Region-Based Convolutional Neural Network object recognition research, a key input for automated surveillance (the US share is 11%) [465]. Of course, publication rates alone do not rule out responsible use of this research.

*3.7.3   Turning Principles into Action.* One concrete product of research on AI ethics is the new norm in corporate AI products of providing "model cards": descriptions of the training process, the provenance of the data, performance limitations, and ethical considerations [388]. Similarly, requirements such as those at the US National Science Foundation for researchers to document the "broader impacts" of their work encourage researchers to think about downstream consequences of their work at its inception [409]. Unfortunately, it is difficult to quantify the effect of these requirements on AI research.

In the West, AI workers play a key role in promoting ethical use [411]. The global talent shortage in the field gives researchers and engineers power over project selection, as demonstrated by the successful walkout over military applications at Google [505, 411, 52]. Protection for whistleblowers is thus a key way for governments and industry bodies to promote responsible AI [126]; such protections are simply not available in many countries, and are not always honored even when they are mandated by law.

As much as AI poses serious risks, there is a precedent for data-driven mechanisms reducing discrimination, and even greater potential [356]. The original FICO credit score—a pioneering loan approval system based on metrics rather than bank managers' judgment—helped expand opportunities Black and African American entrepreneurs to obtain financing in the United States [422]. In an example of future promise, one report concludes that current ML systems could substantially assist each of the UN Sustainable Development Goals [240].

To make systems perform properly for underrepresented groups, we need predictive data about such groups [438]. Federated learning promises to allow such data to be collected while respecting privacy [281], but more representative data needs to be identified.

Much attention has been devoted to AI misinformation [143]. Simultaneously, a line of AI systems have shown promise in *correcting* misinformation and catching bad scholarship [456, 454, 143]. This could be a scalable way to address the rise of generative models. For instance, the Meta AI system Sphere scans citations, looks up the original sources, and detects cases where the source does not support the claim [456].

## 3.8   Hard Problem #8: By 2050, we will have solved AI-related **risks around its use and misuse, competition, cooperation, and coordination between countries and other key actors**.

AI is likely to transform international security, the broader competitive landscape between states, the power and roles of states relative to other actors, and the conditions or foundations of global cooperation or coordination. These geopolitical challenges will be particularly difficult to address: "few eras have faced a strategic and technological challenge so complex [as AI] and with so little consensus about either the nature of the challenge or even the vocabulary necessary for discussing it." [296]. The history of cyberwarfare and drone warfare shows that it can be difficult to anticipate the impacts of emerging technologies, or even to assess them accurately *after* deployment [250]. This lack of consensus about the exact impacts, coupled with the high political stakes around AI, will likely impede effective coordination. Indeed, one hurdle to managing AI's geopolitical impacts is that addressing them is a collective action problem [412] that entails structural risks, such that there is no apparent entity whose behavior could redress the harm. As a result, adverse impacts of the technology will likely be diffuse and uncertain, and effective responses will be delayed [127].

Figure 11 predicts the effect of geopolitical applications of AI on different elements of the international order. Only the use of AI for information gathering (reducing uncertainty and validating counterparty claims about capabilities) is thought likely to reduce tension; the authors find that other effects are likely to reduce stability, and to erode the compliance of military operations with humanitarian principles.

*3.8.1 Destabilizing military conflicts.* AI has the potential to destabilize international relations in the near term. While militaries face unexpected challenges in procuring AI technologies and adapting them to their operational needs [33, 572], recent years have seen growing investment in AI by G7 countries [226], with Libya and Ukraine providing opportunities to demonstrate these systems at work [360, 367]—to much outspoken opposition from civil society and public stakeholders [52, 36]. The advancement of AI capabilities in the military and intelligence realm may lead to loss of strategic stability, a disturbance of the foundational tenets of nuclear deterrence [296, 251, 585, 367, 32], and new avenues towards unintended escalation [275]. Off the battlefield, AI makes cyber-threats cheaper, more scalable, and potentially more dangerous [80, 172, 276].

The growing availability of AI means that these systems are likely to proliferate to non-state actors. This includes AI-enabled systems which could facilitate terror attacks and assassinations [59]. Although truly catastrophic terror attacks using swarms of small drones may encounter practical constraints in the near term [367], it is increasingly recognized that AI systems for tasks such as image recognition need not perform flawlessly to be attractive to non-state insurgent groups seeking to level the playing field against state militaries [318, 376].

Addressing these risks is complicated by the fact that leading nations have generally shown a lack of interest in regulating military AI [150, 360]. While confidence-building measures, norm-setting and ultimately treaties have helped limit the destabilizing impact of chemical, biological and nuclear weapons, AI is rapidly finding revenue streams that eluded the nuclear industry, with a speed and magnitude of scaling without precedent in other industries.

*3.8.2 Destabilizing other international interactions.* The geopolitical impacts of AI are far broader than military use. New technologies often significantly disrupt trade and other aspects of international relations, altering states' preferences and interests, redistributing power, and creating new ways to exercise hard and soft power [160]. This is especially true for "general-purpose technologies" like AI [148, 193, 585]. Traditional geopolitical analysis focusing on state actors may become misleading as private companies play an ever more important role in significant AI. Likewise, established concepts like the "national interest" may be complicated by transnational networks of firms and individuals [149].

Another problem lies in managing the growing geopolitical competition over AI. The common description of these trends as an "arms race" is inaccurate, since only a minority of the capabilities pursued or invested in are entirely military in nature, and arms, once created, do not lead to increased productivity [495]. The trends can be more accurately described as a winner-take-most "virtuous circle" with strong network effects. Success in such a competition depends on access to key AI inputs, such as data, hardware, software and talent [326], as well as different states' efforts to deny one another such access through strategic decoupling [553] and the maintenance of "chokepoints" [44] in the global AI value chain.

At present, the US and China lead the AI field [**index2024**]. The technological decoupling between China and the United States is continuing, with China working towards self-sufficiency in high-tech products [47]—for example, in 2022 the Chinese government mandated that all computers used by state organizations must be produced within China. [326]. China's pursuit of AI arises not only from domestic regulatory needs, but from the desire for global agenda-setting power [100]. While leading Chinese private companies such as Alibaba, Huawei, and Baidu are now considered "national champions" and play some role in agenda-setting, their influence is limited compared to their US counterparts.

Strategic competition over AI is driving growing regulatory fragmentation between the US, China, and Europe [481, 177]. While the US and China lead in terms of technical AI capabilities, the European Union leads in AI regulation [523, 328, 163, 326]. Differing regulatory regimes are likely to cause conflict not just between military rivals, but between the individual countries that make up the Eastern and Western blocs, as corporations take advantage of opportunities for regulatory arbitrage.

To date, the US approach to AI governance has been largely hands–off [326, 482], fostering innovation by reducing regulatory burdens [561]. While the US has implemented some important initiatives such as the AI Bill of Rights [66], its overall laissez-faire approach has resulted in less global influence than the framework developed by the European Union. Instead, the US has focused on industrial policy, such as the CHIPS and Science Acts [326, 453, 18], to strengthen its AI supply chain—increasingly to the detriment of its European allies. Lastly, the international nature of AI ecosystems and digital platforms renders exclusively national regulation expensive and suboptimal [498].

*3.8.3 Navigating AI turbulence.* To carry a stable world through to the year 2050, the geopolitical risks associated with AI will need to be addressed, just as in the last century we found ways to address the risk and instability created by nuclear weapons.

AI is likely to play a role. Some AI tools, including translation, information aggregation and monitoring tools, are likely to improve information flow and ultimately decrease conflict—especially given their potential to help smaller actors participate in global dialogue on a more even footing [137, 365].

## 3.9 Hard Problem #9: By 2050, we will have solved the **adaptation, co-evolution and resiliency of institutions and social infrastructure** to keep up with and harness AI progress for the benefit of society.

> "This lack of [AI governance] standards makes it both more challenging to deploy systems, as developers may need to determine their own policies for deployment, and it also makes deployments inherently risky, as there's less shared knowledge about what 'safe' deployments look like. We are, in a sense, building the plane as it is taking off."
>
> – Ganguli et al. [192]

The pressures placed by AI on social infrastructure (for example governments, universities, corporations, publishers) are likely to increase as AI capabilities continue to grow, proliferate, and outstrip control. HP#9 focuses on domestic governance: we consider explicit governmental and industry mechanisms for AI assurance, alignment, social responsibility, and the adaptation of social institutions to the coming disruption.

Stahl [533] divides governance proposals into three main categories: policy-level proposals, organizational responses, and guidance for individuals. We focus on the first two. (Policy-level proposals are also called "hard law," while the other two categories fall under "soft law.")

While AI governance has grown significantly in recent years [514, 19]), the field is still finding its place between public policy, political science, international relations, security studies, jurisprudence, and other existing structures.

Figure 12 depicts some mechanisms by which AI governance is produced. Such mechanisms cover regulation, industry standards, ethics principles, and coordination protocols between stakeholders. Although regulations and standards serve as potent instruments in governing AI, their efficacy in directing AI towards socially advantageous outcomes may be limited without the integration of ethical considerations and the capacity to *apply* such ethical principles, [183, 140], as well as research into risks that are not yet within the "Overton window" of political concern [127].

In governance papers published in 2010–2022, the most common academic background was computer science and IT, with 52% of researchers coming from other fields [19].

Up to 2023, contributions to AI governance focused on on individual problems [537, 478] and attempts to lay foundations for the field through prospective research agendas and advocacy [125, 128, 588]. In 2023, this work bore fruit in the form of a major international summit [562], Congressional and House hearings [565], consensus declarations [564], and new, well-resourced government bodies with responsibility for the oversight and evaluation of AI risks [563, 418].

A major problem with surveying governance research is that the public record is heavily filtered. Jack Clark, former policy director at OpenAI, notes candidly that "A surprisingly large amount of AI policy is illegible, because mostly [only] the PR-friendly stuff gets published, and many of the smartest people working in AI policy circulate all their stuff privately" [109].

Some problems with AI regulation are shared with other emerging technologies: consider information asymmetries between manufacturers and regulators, policy uncertainty and the political cycle, structural power dynamics, and policy errors [546]. The resulting "Collingridge dilemma" is the observation that regulators attempting to prevent harm from new technologies must create norms and laws before it is even possible to fully understand the potential impact of the technology, or their regulations on it [200, 514]. Regulators face a trade-off between proactive regulation (to minimize risks but hinder innovation) and a more laissez-faire approach (that would incentivize AI development, but could drastically increase potential risks).

The dilemma is common to many emerging technologies, but the field of AI governance is extreme in that its potential risks include proactive and adversarial dynamics not found in other technologies: AI is a technology which can misuse itself [126, 125, 92]. In addition to directly catastrophic scenarios such as misaligned power-seeking AI [92], Dafoe identifies four main sources of risk: (1) robust totalitarianism; (2) preventive, inadvertent, or runaway nuclear war; (3) powerful AI systems not fully aligned with human values; and (4) systematic value erosion from competition [127].

Gasser and Almeida categorize approaches to AI governance as a series of layers on top of the technology:

(1) *the technical layer*, which seeks to understand AI technologies, leading to context-specific mechanisms for different technologies and different levels of intervention (technical, organizational, policy);
(2) *the responsibility layer*, in which all the societal, ethical, and regulatory impacts of AI are considered. The aim is to help guide policy by understanding the wider societal, ethical, and legal challenges;
(3) *the regulation layer*, in which the specific subjects addressed by AI regulation and coordination are defined;
(4) *the public policy layer*, including the implementation of hard and soft governance mechanisms (social and legal norms, regulation and legislation, ethical principles and codes of conduct, as well as practices such as data management tools, standards, and certifications); policy implementation should take into account the various contexts and levels of implementation in the technical layer and involve various forms of cooperation between different actors and stakeholders;
(5) *the collaborative layer*, in which stakeholder goals and conflicting interests are balanced, building trust, shared values, and motivation among different stakeholders [589].

A leading example of a policy which spans these layers is "compute governance", by which the provenance of high-end chips is tracked and controlled [291, 295]. Other tools include procurement regulation [155], "soft law" [369], ethical reviews and impact assessments at conferences [313], research release strategies [509], pre-publication impact assessments [459], standards-setting organizations (e.g. ISO, IEEE, ESOs, and NIST) [417, 106], third-party certification schemes [107], criminal law [293], human rights law [355], ethics oversight boards (on the model of the Meta Oversight Board [234]), permit programs [566], cooperative policies between AI labs [30], software and hardware mechanisms that allow actors to make verifiable claims about their AI systems [81], competition law [231], and perhaps new instruments like private regulatory markets [110].

### 3.10 Hard Problem #10: By 2050, we will have solved **what it means to be human in the age of AI**.

AI poses personal and philosophical challenges. Say an AI does your job better than you; setting aside the economic concerns, what does this mean for you as a moral agent, as a citizen, and as a human being?

Figure 13 presents John Danaher's list of philosophical and personal problems that would arise if AI and other software greatly reduced the economic and social importance of human labor and creativity [131].

*3.10.1 Does AI mean that humans won't matter?* A common goal in life is to make a difference, but the conceivable full automation of human work would prevent this. Danaher calls this the *severance problem*: automating human activity *severs* the connection between human effort and the world improving (counterfactually, such that the world would be worse if we did not act) [131]. This problem also covers the loss of positional goods, which rely on us being the "best" at something (perhaps as a species, rather than individually) [77]. Moreover, the problem is only partially mitigated by limiting machine entry into certain fields—for example by preventing bipedal robots from entering into races with humans—since the essence of the problem is that you *could* be replaced even if you are not.

The severance problem is related to the worry that AI excellence will inspire passivity in individuals, and so stagnation in society [421, 132]—a kind of bystander effect for values. This is particularly worrying from the perspective of political and moral agency. Democratic societies require a certain level of voluntary action, activism, oversight, and other forms of participation to function in a stable fashion [129]. Similarly, moral progress usually depends on the actions of the unusual thinkers and 'moral entrepreneurs' [134] who start the movements that lead to shifting attitudes and laws [567].

Language is the major channel of social organization. Tobias Rees argues that language models thus precipitate a particular crisis because our culture (or ontology) has assumed that humans are the only agents which use language, and that we are unprepared for a world in which most text is AI-generated [473].

AI could make it drastically harder to understand society and each person's role in it [131]. For example, systems collect data on us and make decisions about us—but if these processes are inside a private company (or a classified government function), we will struggle to discover and establish an explanation for decisions made about us [501, 117]. This could greatly intensify the old problem of bureaucracy: diffused responsibility, without possibility of appeal or transparency [268, 361, 373]. Inscrutable systems such as large neural networks will remove the possibility of understanding how specific decisions were made, even in the presence of strong legal protections and compliant organizations [86]. Together, these effects serve to make society's operations opaque to a given person, and so make it harder for them to exercise control over their own life. This arguably also threatens their dignity [131].

Finally, the ongoing spread of AI surveillance could also threaten freedom of association and protest, even in democracies [547]. AI and automation will make it cheaper to exercise various forms of social control [179, 590]. This can be achieved in a (relatively) liberal fashion through 'nudging', subtly altering the choices made most salient to the user [72]; adapting this to best influence *each user* differently has been called "hypernudging" [324]. Digital policing and omnipresent surveillance is another, much less subtle, way to have your freedom limited—with its increased scope and scalability allowing for new forms of overreach [179].

If your decisions are covertly or overtly influenced, making it harder for you to exercise control over your own life, is this a threat to your dignity [131] and what it means to be human? Or will this simply become the 'new normal', a default which does not often rise to conscious attention?

Besides the potential loss of broader flourishing, we might worry that loss of attention could further reinforce a fall in social participation, as discussed below.

*3.10.2  Will humans choose to value things that don't matter?* Alternatively, humans may chose to retain their distinctiveness from AI by purposely excluding AI from particular activities, by analogy with a wealthy family that prefers to cook a meal themselves, rather than have one cooked by a professional chef.

Consider sport, one arena in which human excellence and meaning-making has not suffered much from the existence of superhuman machines [272]. Chess engines have been superhuman for 25 years [89], but human chess retains an audience many thousands of times larger than machine chess—and by many measures, interest in chess played by humans is higher than it has ever been [550]. This is some indication that human-human competition may be both preferable and sufficient for meaning.

The drama that grips audiences also seems to stem from human performers [261]: however, whether humans will prefer live-action humans in pre-recorded video when actors are indistinguishable from artificial replacements is an open question. One point in favor is that live performance of all kinds is often regarded as a superior good to pre-recorded or generated art, even all-time great recordings [543, 320]. Live shows generally command a cost premium. Further, if the meaningfulness of these activities survives machine excellence then it seems likely that the goods associated with connoisseurship and participation in the related communities will also survive [368, 142, 555]. Human performance might then still generate some form of the crucial goods of excellence, mattering, community, and social recognition [201]—even if it is indistinguishable from AI output to most observers.

*3.10.3  Meaning after AI.* Is our last paragraph wishful thinking by "meat machines"[541] on our way to being replaced by superior content creators? Like the current economy, the coming AI economy may default to producing an unending stream of addictive or distracting entertainment—perhaps with psychological consequences far beyond current claims about addictive social media [252, 93]. The philosophical problem with addiction (or extreme distraction) is that it replaces broad life pursuits with one-dimensional behavior which likely misses much of what others consider to be valuable (the addicted individual having lost the ability to behave as a rational actor). Highly addictive AI-developed media might prevent us from realizing indirect goods which require delayed gratification and investment, especially social goods [180]. This view presumes pluralism, in that there are many valuable things, and it is important to achieve a good range of them [547, 184]. (Philosophers often assume that addiction leads to the total loss of self-control, and so dignity, but this is contested [184].) Another problem is that automated entertainment will reflect "simple engagement metrics rather than a harder-to-measure combination of societal and consumer well-being" [68]—and thus provide us with unusually shallow kinds of pleasure.

Given this background, it seems odd to suggest that AI could help resolve the subjective and philosophical problems it causes. One route involves helping us understand ourselves [209]: if the scientific promise of AI is met, then we will all gain the good of better understanding the world, society, and the brain. AI might also provide an unprecedented laboratory for simulating and testing hypotheses about intelligence, creativity, cultural evolution, and values [473, 31, 27].

## 4  Are the Hard Problems wicked problems?

Rittel and Webber coined the term *wicked problems* to describe social problems that *by nature* cannot be solved [479]. Wicked problems defy solution due to their entanglement of facts and values; their lack of a conclusive criterion; their preclusion of any opportunity to learn by trial-and-error; their numerous incompatible explanations; and to their perverse links to other problems. For example, bringing fresh water to buildings and removing waste through city-scale plumbing and sewer infrastructures is not a wicked problem, because approaches for building water delivery and sewage are well understood, and near-optimal approaches can be readily designed. On the other hand, designing an urban mass-transit system requires weighing many more choices, with each potential decision having many potentially positive and negative consequences.

The wicked problem framework has been widely used in the decades since it was formulated. As Hou, Li, and Song write in a review of 800 academic publications that used the wicked problem framework, "Because fragmented but interconnected social challenges coexist, wicked problems call for cooperation among different disciplines" [253].

Surveying the above research, it may be tempting to conclude that our Hard Problems are wicked problems, and so move on to some less question—but we claim that suitably realistic and precise versions of the Hard Problems are not wicked. After forty years of work using the framework, "Most authors hold that wicked problems are made up of multiple internal characteristics, all of which can be divided on different scales from docile to intractable, as differences between facts and values [in] combination," Hou, Li, and Song continue; what makes a problem wicked, rather than tame, is that it can "cross the boundaries between countries, policy domains, organizations, and scientific disciplines" with "a combination of complexity, diversity and uncertainty" [253].

Making progress on a wicked problem requires distinguishing aspects of the problem that are based on *facts* from those that are based on *values*. Problems that are largely about facts should have near-optimal solutions, given agreement on a loss function for mapping multiple parameters onto some scale. Values are inherently more complex than facts, both a result of their philosophical foundations and the difficulty of quantifying them.

The Hard Problems fit some of the wickedness criteria—for example, there is no 'stopping rule' for understanding or redefining what it means to be human (HP#10). We may not have an answer for what it means to be human in the age of AI, but we have a working understanding of the more general problem via the history of philosophy.

When overcoming the "scientific and technological limitations in current AI" (HP#1) or enabling "game-changing contributions" (HP#4), we do have "conventionalized criteria for objectively deciding whether the offered solution" [479] does or does not address today's AI limitations or create new capabilities, violating one of the criteria for wickedness. These solutions (if they are found) will also be testable, which violates one of the worst features of a wicked problem.

The Hard Problems that directly invoke human values (e.g., HP#3, HP#6–9) are better candidates for wicked problems: they lack stopping rules and the purported solutions are not testable. With such problems, solutions are not judged on technical correctness, but on their fit to values. For instance, a technology that supports responsible democratic control by an informed electorate will only make progress on HP#9 for users who prefer democracy. For those who prefer a totalitarian system in which artificial intelligence surveillance protects a ruling elite against accountability [90, 607, 147], very different breakthroughs will be seen as progress.

By arguing that the Hard Problems are not necessarily wicked, we hold open the possibility that some or all of the problems might actually be solved—or at least, that we might be closer to solving them in 2050 than today.

## 5  Outlook

Between 2012 and 2022, the publication rate in technical AI doubled, reaching 240,000 papers per year [**index2024**], more than the entire field of physics [517]. As a result, the most intensely researched and well-resourced areas in AI are HP#1 and HP#4—the accelerator pedal.

The broad vision of the AI2050 program is that AI researchers, funding agencies, and society at large can make decisions now and in the following years that will result in widespread beneficial impact. Finding answers to the Hard Problems will clearly be a challenge, but we see no other choice if humans are to coexist with our technologies. For all these problems, one clear condition of a successful 2050 scenario is that we begin work on them now.

## Acknowledgments

This literature review was made possible in part by grant G-22-63887 of the Eric and Wendy Schmidt Fund for Strategic Innovation, and by the Schelling Residency. We thank Sanjeev Arora, Mike Belinsky, Jan Brauner, Dan Carey, Connor Coley, Andis Draguns, Owain Evans, Basil Halperin, Jose Hernandez-Orallo, Robert Kirk, James Lucassen, Matthijs Maas, Sören Mindermann, Hugh Panton, Javier Prieto, Yonadav Shavit, Karina Vold, Sophia

Wisdom, Junfeng Yang, John Zerilli, Huan Zhang, and our many anonymous reviewers for their comments and suggested improvements to the paper. We thank Calum Leslie for editing work. Marcus Gemzoe-Winding and Anekdote Studio assisted with our visualizations. Eric Schmidt and James Manyika contributed the original list of hard problems.

# FIGURES AND TABLES

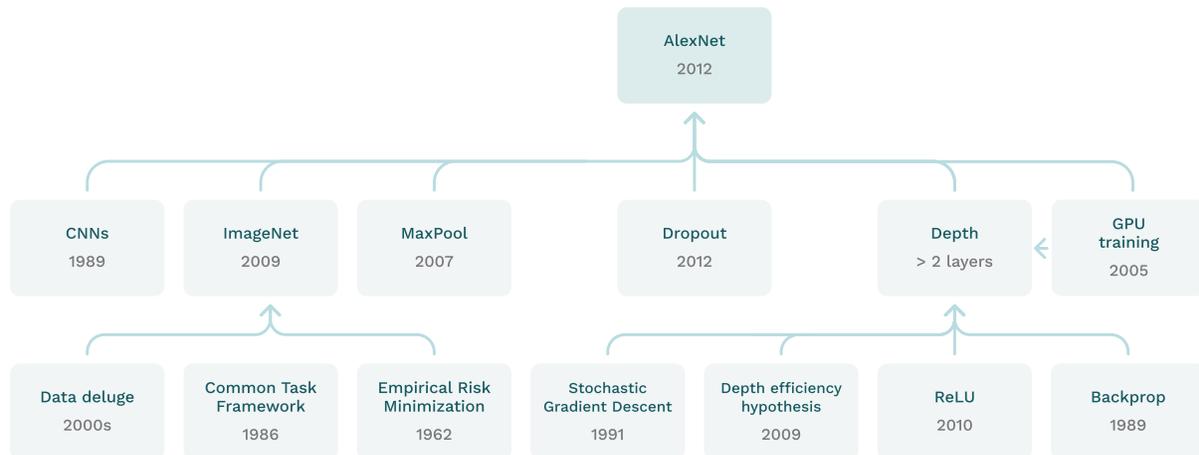

Fig. 1. The path to the DL breakthrough in computer vision (2010–2012), using [108, 310] as an exemplar, following [58, 55, 207]. No part of this graph is a necessary condition (besides massively increased computation, here represented by GPUs, and some stabilizing activation function); see [28] for a very different minimal path. The "data deluge" is the internet-driven availability of data on more or less everything [53]. The "Common Task Framework" is competitive benchmarking: the use of fixed datasets and objectives, with held-out test data, plus a culture of open competition (Liberman [348] dates it to 1986, but see also [470]). ImageNet is the most famous example of a common task: for its time, it was an extremely large and diverse image classification dataset [144]. Empirical risk minimization is the fundamental method of machine learning: creating a training set and a test set and using training performance as an estimate of true performance [241]. Convolutional neural networks (CNNs) were the original network architecture employed during the DL liftoff [330]. Max pooling (MaxPool) is an approach to reducing the state of a convolutional layer ([468], following [594]). Dropout prevents overfitting by randomly deleting connections between units ([244] following [527]). Stochastic gradient descent is a remarkably efficient general optimizer ([71], following [480]). The "depth efficiency hypothesis" was the long-standing belief that adding layers should improve the statistical efficiency of networks, exemplified in [55, 57] and later proven in [444]. Graphical processing units (GPUs) sped up training by an order of magnitude or more [536]; reuse of this pre-existing hardware represented a lucky ticket in the "hardware lottery" [249], shifting future hardware investments towards DL operations. Backpropagation is the reigning method of assigning "credit" to particular weights, enabling gradient-based learning in neural networks ([330, 489] following pioneering work by [586, 159, 350]. The rectified linear unit (ReLU) is an activation function which avoids the vanishing/exploding gradient problem ([403], following [187]). "Depth" is the number of layers between input and output in the network; early systems counted two hidden layers as "deep" [243, 58] while AlexNet had eight—and, ultimately more importantly, 60 million parameters [310]. (AlexNet was not the first success in computer vision or neural networks; eight years before AlexNet, Kussul and Baidyk achieved better than 99% accuracy in handwriting recognition [316].)

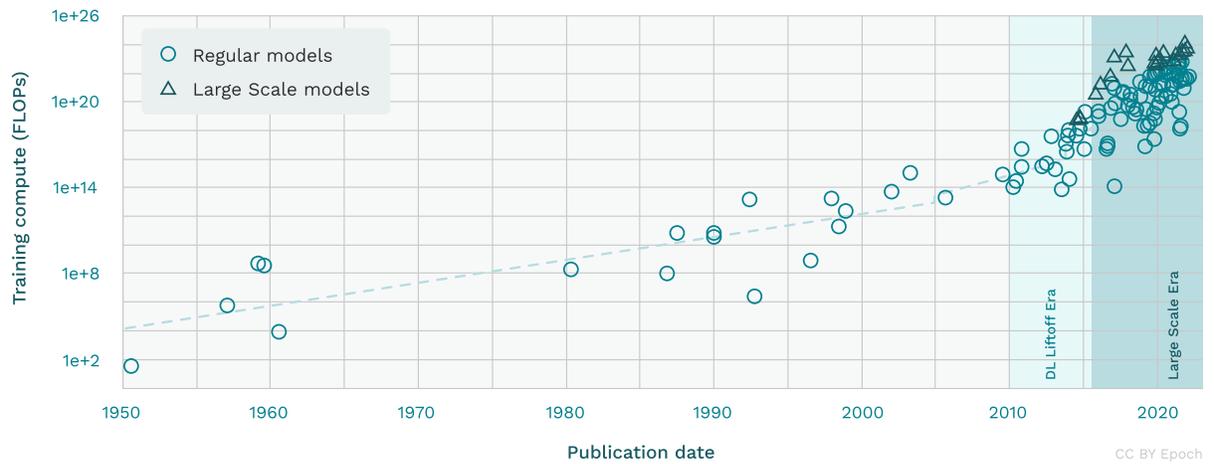

Fig. 2. Timeline of computation required (in floating point operations, FLOPs) to train leading machine learning systems (n=121): note the exponential increase. A "milestone" model is one that has "an explicit learning component, showcases experimental results, and advances the state-of-the-art and at least one notability criterion" (>1000 citations, retrospective historical importance, deployed in a notable context) [504]. Where the total compute was not given by the paper, the Epoch team estimated it using methods from [21]. Using a separate fit for models more than three-quarters of a standard deviation above that year's mean compute, they obtain three plausible trends, somewhat robust to the date or size thresholds. (Note the huge practical significance of a slightly different slope on an exponential scale.) The graph depicts three eras with different shading: the pre-deep learning period 1950–2010; the deep learning liftoff period 2010–2016; and the large-scale period 2016–2022. Each ○ circle is one milestone system, n=91; triangles denote a separate cluster of anomalously large models, n=16. The graph shows that the exponential increase in demand for computation per model accelerated around 2010, when deep learning became popular; it also shows the introduction of anomalously large models in 2016. Adapted with permission from [504].

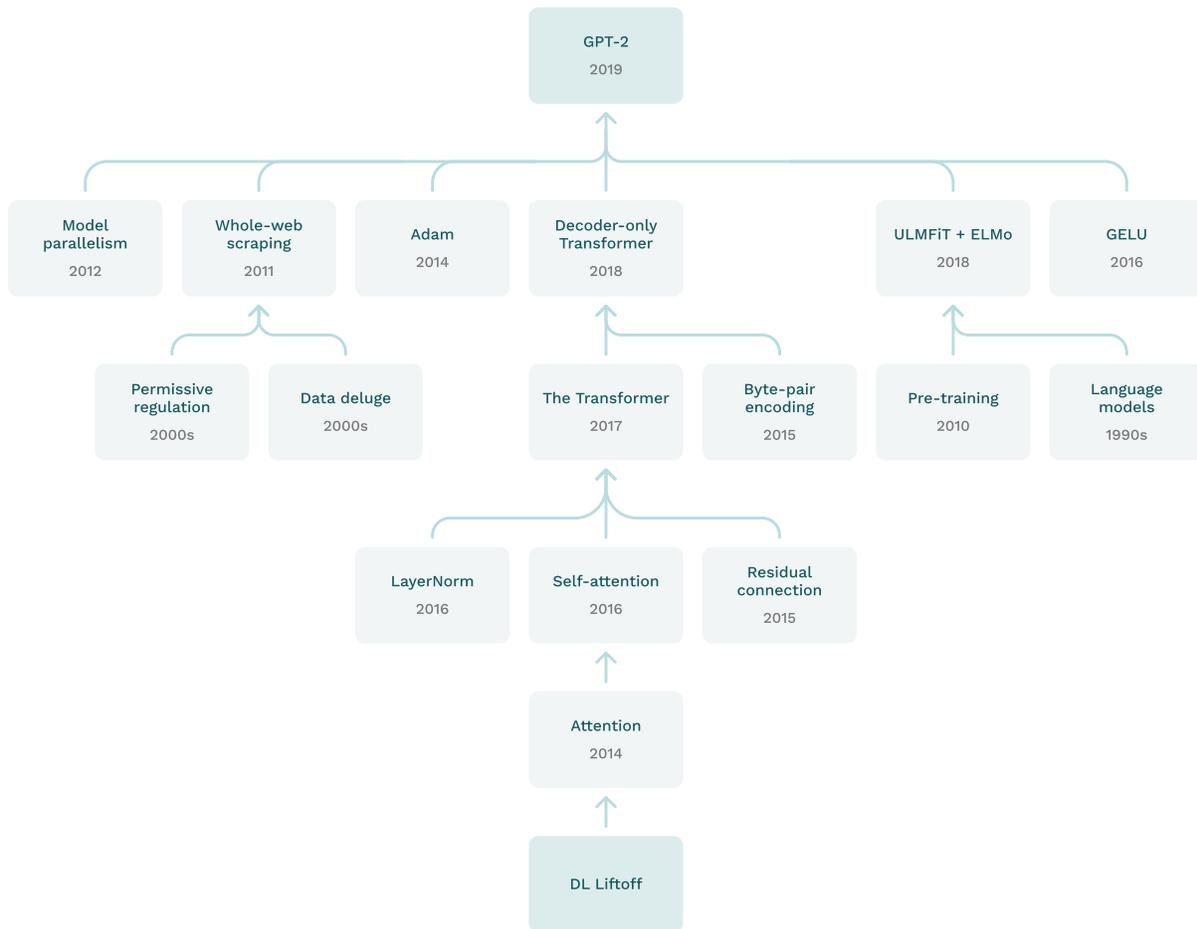

Fig. 3. The path to the large scale era in deep learning (2016–present), using GPT-2 as an exemplar. The most important node in this graph is pre-training, read as a stand-in for massively scaled parameter counts. Adam (after "adaptive moment estimation") is a robust replacement for stochastic gradient descent which maintains and adapts many separate learning rates [294]. Unsupervised pretraining (preparing a neural network for later tasks by learning generally useful features from unlabeled data) gained new relevance when used for sequence modelling [130, 146] (rather than its original purpose, stable network initialization [168]). Large language models depend on web scraping at the scale of the entire internet [580], which in turn depends on relaxed attitudes to copyright on the internet. ULMFiT and ELMo pioneered unsupervised training that produced transferable skills and embeddings for NLP [254, 452, 461]. GELU is an activation function with better implicit regularization [236]. Attention weights the parts of several input sequences in proportion to their relevance to the decoding at the current training step [37]; self-attention, o n the other hand, weights parts of a single input sequence, allowing us to infer dependencies [99]. Byte-pair encoding tokenizes input text into a code suitable for learning subword representations ([503], following [190]). The Transformer is a network built from many layers of attention ("multi-headed attention"), plus a positional encoding separate from the text (essentially an accumulator indicating word order) [570]. The decoder-only form of the Transformer simplifies the architecture and improves zero-shot generalization at the cost of being unidirectional [353]; many state-of-the-art models are instances [579]. Layer normalization stabilizes training by using summary statistics of inputs to an entire layer of the network [34]. Model parallelism splits a model across training machines (often by layer) for enormous model sizes and parallel speedups [135].

| General practice | Deep Learning Liftoff | Large Scale Era |
|---|---|---|
| Training signal [1] | supervised | self and semi-supervised |
| Task model [2] | discriminative | generative |
| Tasks per system [3] | single-task | multitask |
| Input types [4] | single-modality | multimodal |
| Design specialization [5] | > one architecture per domain | increasingly one architecture |
| Inductive bias | hand-crafted | increasingly, meta-learned |
| Hardware setup [6] | single training machine | distributed training |
| Hardware [7] | consumer GPUs | 'AI accelerators' (ASICs) |
| Initialization | trained from scratch | finetuning pretrained models |
| Connectivity [8] | dense | perhaps increasingly sparse |
| Role of theory | weak guide to tinkering | curve-fit 'laws', asymptotic theory, some guidance |

[1] Supervised learning is the classic statistical task of fitting data labeled with the "ground truth" (output data which is at least roughly correct) [329]. Self-supervised learning uses entirely unlabeled data, avoiding expensive labeling; semi-supervised learning expands a labeled dataset with unlabeled data [596].

[2] Discriminative models draw a decision boundary in data space, while generative models instead learn the distribution of the data; this allows them to generate typical examples, as seen in the GPT series [79, 428] & DALL-E [208].

[3] Single-task learning refers to systems being trained on only one task, as opposed to learning representations useful for multiple tasks.

[4] Single modality systems (trained on only one data type) have been followed by multi-modal systems handling e.g. both text and image data.

[5] Relatively domain-specific architectures like convolutional and recurrent neural networks were followed by the Transformer's state-of-the-art results in language, vision, audio, and more [349].

[6] In the liftoff era models could be trained on a single workstation, but scaling requirements have led to massively distributed training across thousands of machines.

[7] ML hardware largely comprises repurposed consumer-grade GPUs, succeeded (at the top end) by ASIC "accelerators" specialized for ML operations like matrix multiplication [476].

[8] *Connectivity* refers to the degree of connection between units; "dense" (fully-connected) networks could eventually give way to more efficient, sparser models [245].

Fig. 4. Comparison of rough tendencies between the two eras in deep learning research, besides the key difference that training in the Large Scale Era involves more than a billion times more data and compute [504]. See Figure 2 for definitions of the eras.

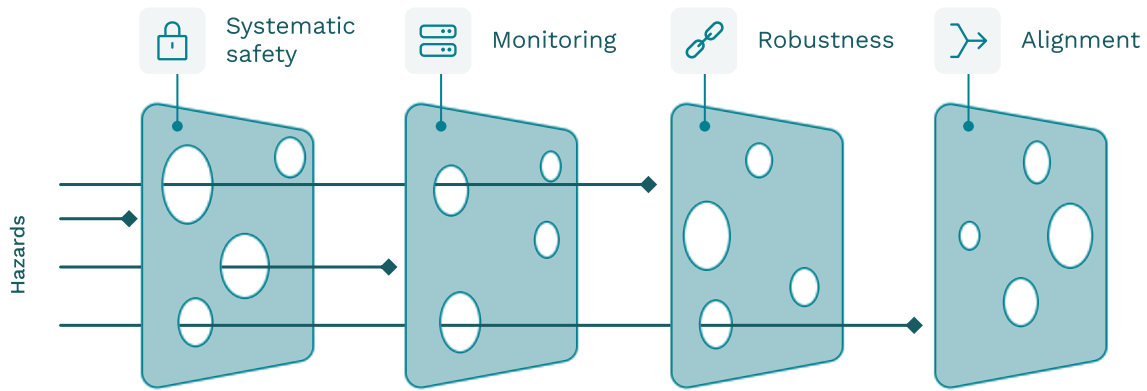

Fig. 5. HP#2: The "Swiss cheese" model of safety engineering as applied to machine learning systems. To reduce the probability of hazards, we can adopt a "defense in depth" approach: a series of mechanisms that each work against some subset of hazards. *Systematic safety* combines a strong safety culture among developers and users with AI systems that detect and block hazards early. *Monitoring* involves gaining greater insight into AI systems themselves, for instance through anomaly detection [237]. *Robustness* measures increase systems' ability to perform gracefully in the presence of extreme events and adversarial action. *Alignment*—ensuring that the goals of advanced systems reflect human values—is the subject of HP#3. Adapted with permission from [238].

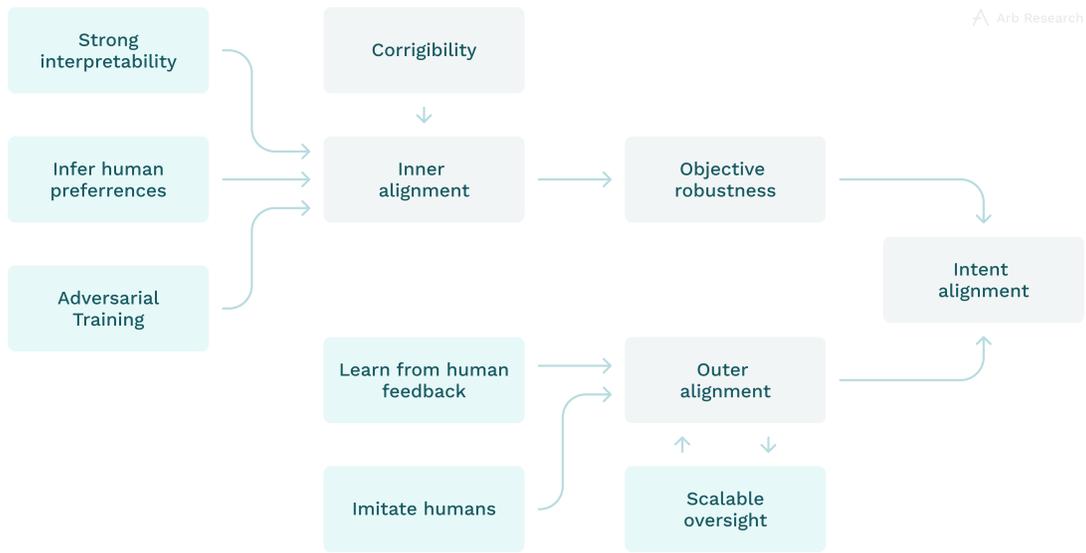

Fig. 6. HP#3: *Alignment approaches* and how they relate to different *alignment milestones.* Intent alignment is making a system try to do what we intend it to. It factors into objective robustness (that the system successfully infers what we intend, even when operating in very different conditions from training) and outer alignment (that the objective given to the system captures important variables and is not merely correlated with the true goal) [260]. Inner alignment (discussed above under "emergent goals") is ensuring that the goal a system *ends up acting under* is safe [260, 337]. A "corrigible" system is one with no incentive to prevent shutdown or modification by users [524, 218]. For a more detailed and up-to-date review, see [332].

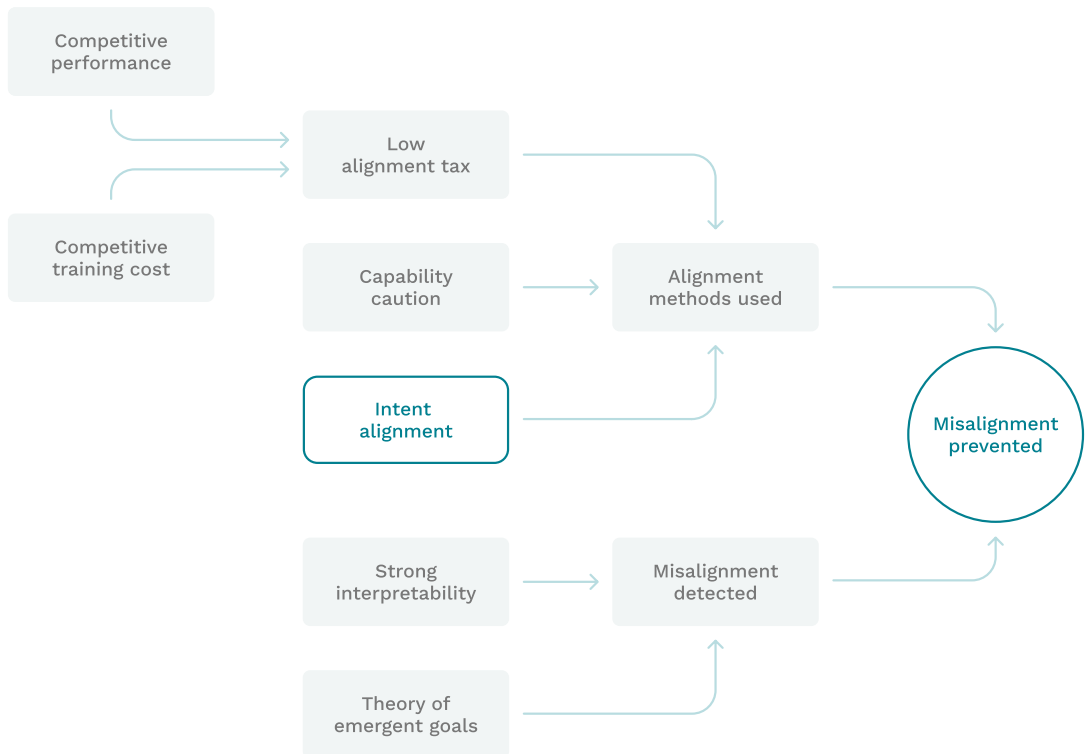

Fig. 7. HP#3: The greater alignment problem: even if methods of intent alignment succeed, we need them to cover powerful systems and to be adopted by a huge variety of actors, which requires the method to have little overhead in training cost or performance (a low "alignment tax"), as well as strong understanding of the risk among AI researchers ("capability caution"). Simultaneously, other misaligned systems need to be detected [70], and alignment problems arising from the perverse interaction of individually-aligned systems solved [158, 121]. Figure inspired by [260, 101].

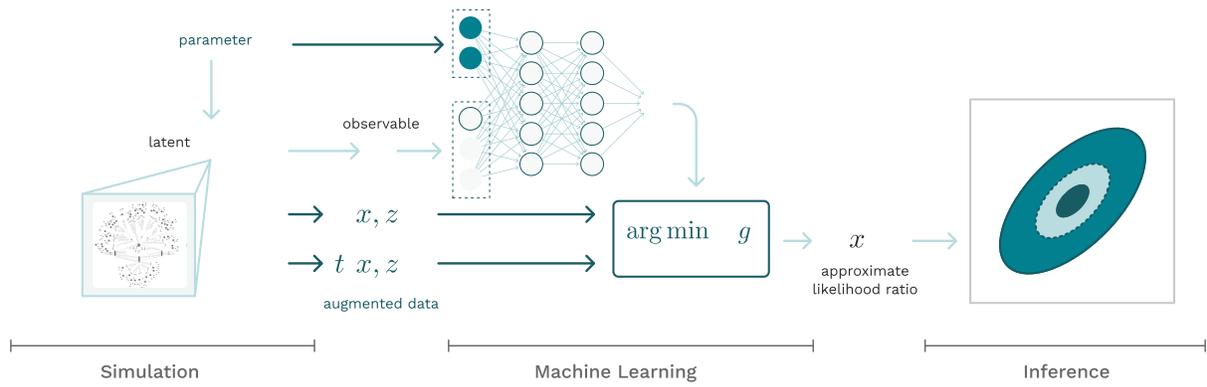

Fig. 8. HP#4: AI creates game-changing opportunities for scientific exploration. ML systems can be trained on high-fidelity but computationally intensive physics simulations. The trained model produces a compact approximation of the target physical system, and can then be used as a proxy for inference. Alternatively, classifiers can be used to characterize actual data from experimental apparatus. Figure adapted from [76].

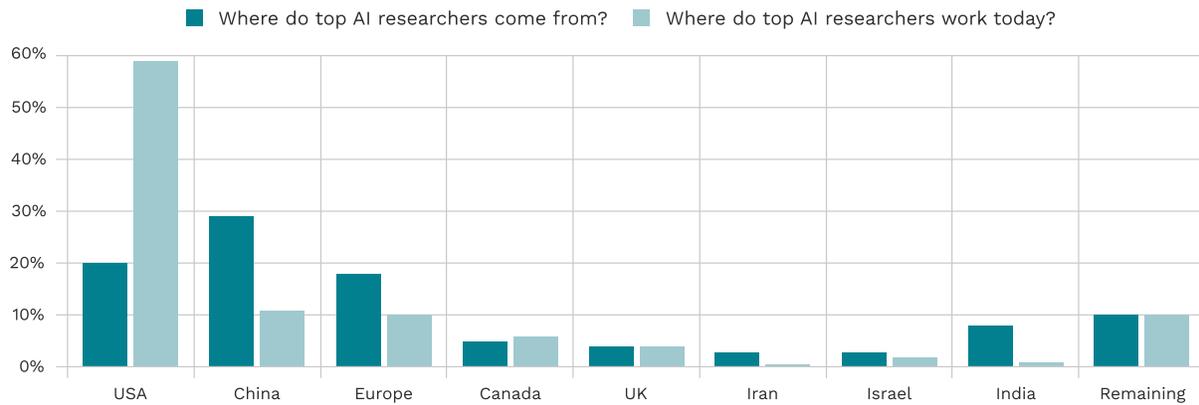

Fig. 9. HP#6: National origin and current location of leading AI researchers (using undergraduate institution as a proxy for origin country, and using publication in the proceedings of one top conference, NeurIPS, as a proxy for top researcher status), 2019. We see that the US is the largest beneficiary by far, gaining 40% of the world's top researchers. The 'Remaining' countries represent more than half of the world's population, but only 10% of its AI researchers. Data from the MacroPolo Global AI Tracker [372].

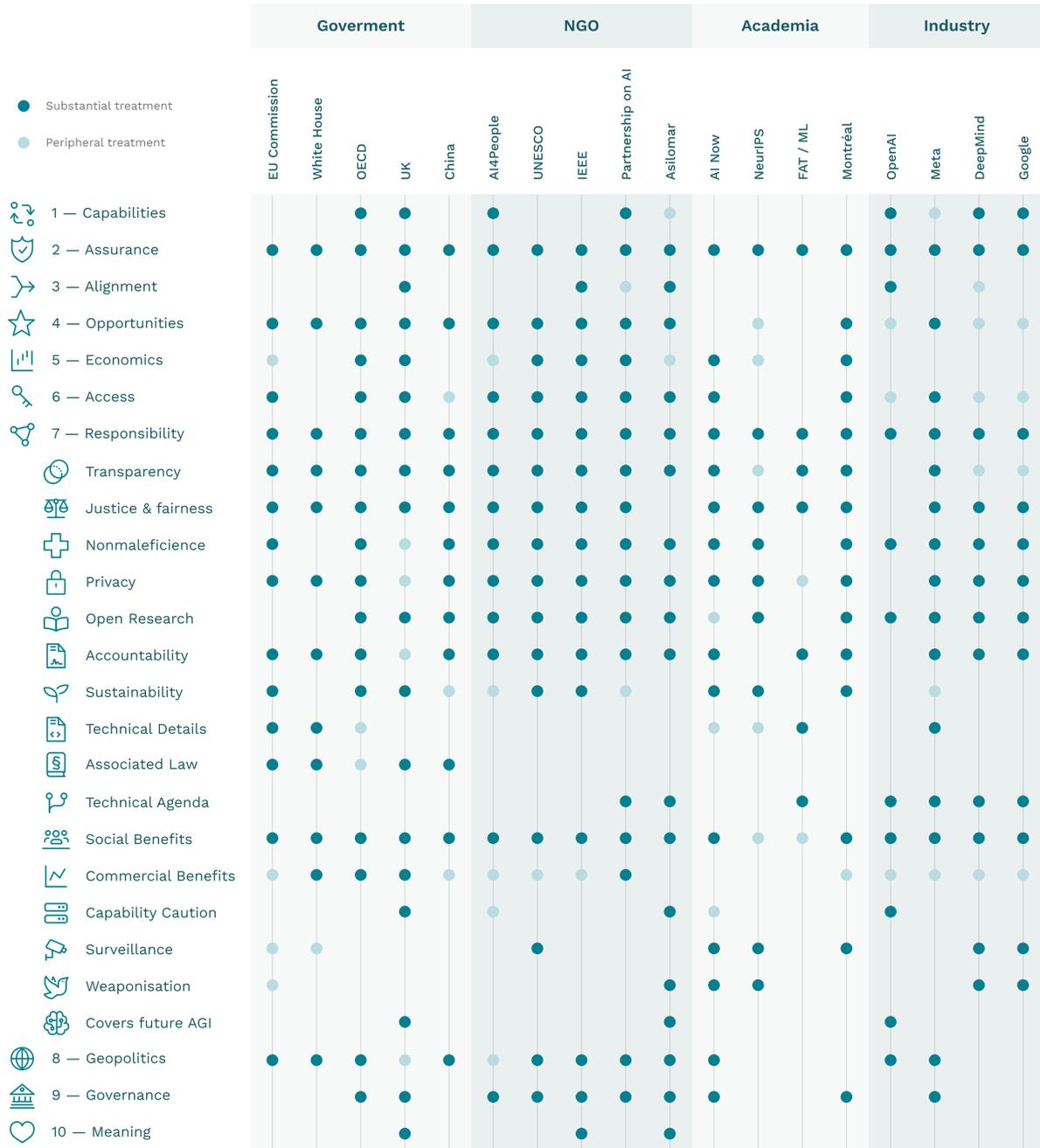

Fig. 10. HP#7: Topics covered in "AI principles" issued by governments, non-governmental organizations, academics, and industry. Citations to the principles and to the associated legislation and technical agendas are collected in Table A.7. Figure inspired by [221].

| Technology | Arms race stability | Crisis stability | Humanitatian principles |
|---|---|---|---|
| AI for C4ISR | Strengthen | Strengthen | Weaken |
| AI for weapons | Weaken | Weaken | Weaken |
| AI for cyber | Weaken | Weaken | Weaken |
| AI for info war | Weaken | Weaken | Weaken |

Fig. 11. HP#8. The effect of AI applications on the international order as expected by researchers at IFSH. "C4ISR" is Command, Control, Communications, Computers, Intelligence, Surveillance, and Reconnaissance: i.e. the entire military and national security information chain. Adapted with permission from [173]
.

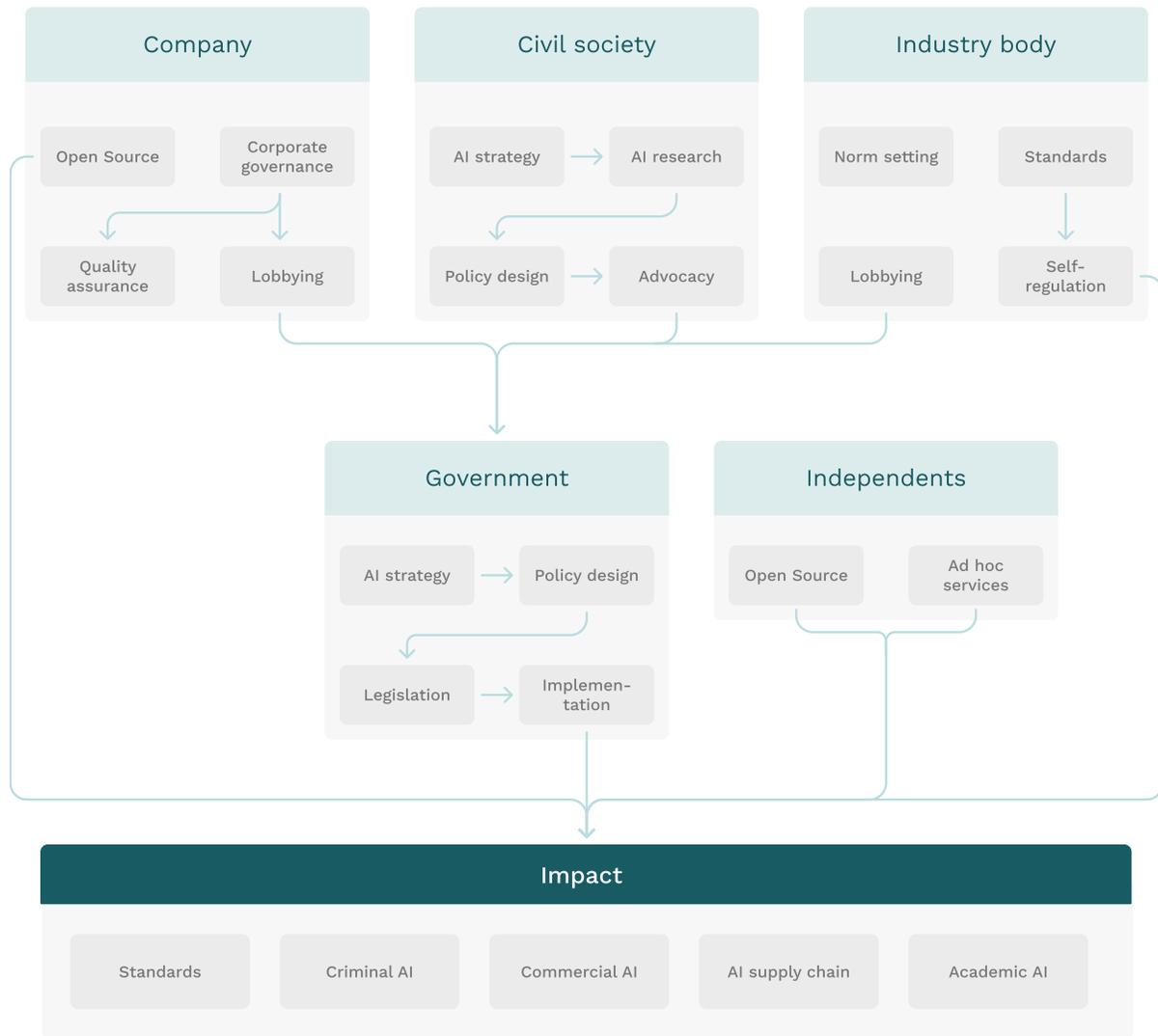

Fig. 12. HP#9: The production of AI governance: actors and mechanisms. 'Civil society' here includes academics, consultants, and think tanks. Independent AI developers are not to be overlooked, since they develop tools and products outside regulatory structures, often using open-sourced weights from industry and academia. Map inspired by [478, 399].

| Problem | Description | What is lost |
|---------|-------------|--------------|
| Severance | By making human activity superfluous, automation severs the counterfactual effect of our actions on the world. | Our sense of achievement and mattering. |
| Attention | Automation is the core of the attention economy, which distracts us from more valuable things. | Intellectual flourishing, creative projects, social goods. |
| Opacity | Systems know things and make decisions about us we do not comprehend because of institutional barriers, lack of knowledge, and the technology being uninterpretable. | Understanding, and so power and dignity. |
| Autonomy | Systems manipulate us by limiting the options we are presented, and through subtle and unsubtle coercion. | Power over oneself, and so dignity. |
| Agency | By making effort seem unnecessary, automation could disincentivize community, goals, duties, and pushing society forward (especially morally). | Sense of being a moral agent. The liberal society. |

Fig. 13.  Potential humanist problems caused by automation technologies including AI. Adapted with permission from [131].

# ELECTRONIC SUPPLEMENT

Note: The entire bibliography is available as a Zotero library at
https://www.zotero.org/groups/4774748/the_hard_problems_bibliography/library.

## A  Additional Material for Each Hard Problem

### A.1  Additional Citations: HP#1

While deep learning led to an improved theory of learning (particularly of the benign overfitting phenomenon [115, 35, 14]), currently we cannot predict whether a DL system will generalize—that is, actually solve the task—when provided with out-of-distribution data [110]. However, humans may not be as sample-efficient as they seem, since each brain is initialized with inductive biases from millions of years of evolutionary experience—our pretraining [171, 93, 149]. In the meantime, ML methods become incrementally more sample efficient year by year, with the occasional dramatic improvement [155, 169, 83, 82, 99].

Other researchers skeptical about the prospects of deep learning alone include [95, 111] and [22].

### A.2  Additional Citations: HP#2

Many authors have noted that there is a significant gap between the average performance of AI systems and the high level of assurance required for deployment in safety-critical environments [76, 38]. Additional references for the potential for AI to create "normal accidents" include [109, 167].

Indeed, many ML systems are now so complex that emergent modes of operation and failure are the *norm* [165, 85].

As Stuart Russell notes: "we are a long way from being able to prove any such theorem [about a behavior being beneficial for humans] for really intelligent machines operating in the real world" [140]. A scalable approach might involve learning (approximate) models of safety properties and the invariants we want our systems to respect.

Technology Readiness Levels[100] are a commonly used scale for describing the maturity of a new technology with regard to its suitability for deployment.

Human oversight, augmented with AI anomaly detection and calibration warnings, holds promise as a component of safety systems [66]. To be viable, such human-in-the-loop systems need to reliably identify cases on which they can safely operate on their own, and only refer cases to human operators which are both impactful and uncertain.[10]

However, extracting guarantees that neural networks will perform within bounds is a daunting task: one classic approach amounts to solving satisfiability problems with as many variables as the network has parameters (i.e. millions or more) [9]. Practical approaches sacrifice exactness and/or completeness [173], and remain limited to small-to-medium parameter counts. Recent work considers increasingly realistic networks, including nonlinear activations and modern architectures [168]. VNN-COMP, a common task competition, finds that hybrid GPU-enabled methods can produce impressive results on networks of up to around 100,000 neurons [13, 172].

Figure 14 shows three approaches to handling the variance of ML performance [114]. Panel (i) shows a system that produces unacceptable performance from time-to-time (as demonstrated by the area to the left of the dashed red line). One approach (ii) adds a failsafe that detects when an AI system might fail, with the result that not many situations are accepted in which the AI's performance is low but acceptable. The approach in panel (iii) avoids opaque AI systems entirely, replacing them with inherently interpretable (or architecturally transparent) systems which users can prove to be safe. Another approach is to reduce the variance of AI performance, either by making the systems themselves more robust, or by improving our understanding of them, as in panel (iv). This figure ignores the cost of producing each system.

Redundancy can be used to produce a reliable system from less reliable components [106]: if a self-driving car has two *independent* stop-sign detectors, each of which is 99.9% reliable, and the driving system stops when either detects a stop sign, then the combined system should be 99.9999% reliable. However, failures in ML perception generally correlate, greatly reducing the effectiveness of this method. Multiple researchers [57, 153] have found that "a wide variety of models with different architectures trained on different subsets of the training data misclassify the same adversarial example" [57].

Another way to improve robustness is to create robust measures of intelligence or generality for AI systems [69, 70], and to use less artificial metrics and more inclusive methods from the natural sciences when studying AI systems [136, 15, 102].

Interpretability is especially important when the loss function used to train the system may not fully represent the interests of the user (or the subject of modeling) [104, 39].

*AI for assurance*    The Eliciting Latent Knowledge (ELK) agenda attempts to develop AI techniques to deduce what a system "believes" about its environment; this line of work could be extremely useful for avoiding unexpected and catastrophic outcomes [24].

AI could be used to detect anomalous inputs, and hence whether the system is operating outside (the training) distribution to an extent which could jeopardize its performance [66, 68].

Adversarial training synthesizes misleading edge cases for other AI systems during their training, making failure less likely when the system being trained encounters such cases in production [174, 79].

*Security*    The above concerns the safety of ML systems: that is, their accident risk. The other major category of hazard is adversarial attack. The essence of an adversarial situation is that the environment adapts to your countermeasures: the noise affecting your system can thus be crafted to be maximally perverse [51]. End users, data subjects or third parties may attempt to manipulate AI systems into making decisions and predictions which benefit them or harm others [20]. This serves to create a second mode in the performance distribution, with an unusually high probability of unusually bad outcomes.

One security measure is multi-criteria evaluation: system performance should be assessed with more than one metric, and a threshold should be met on all metrics [102]. (For instance, a language model may perform exceptionally well at holding a realistic conversation, while simultaneously leaking the personal information used to train it to malicious third parties [132] or producing offensive output.)

Interpretability and simplicity make systems easier to attack, even as they improve worst-case outcomes [104, 52]. Simple, deterministic systems are easier for attackers to understand and thus abuse. Tools which help us find (and fix) adversarial examples will also help attackers find (and thus exploit) adversarial examples [158, 163]. It is thus reasonable to assume that such attacks will become more feasible [20], not least through the use of data poisoning [51] and neural backdoors [56].

In addition to being constructed in the typical way, we can derive "architecturally" [74] interpretable models by training a complex model and then paring it down. Mechanistic interpretability investigates the circuits created during training and deduces what they are doing, ideally to the level of the exact algorithm it has learned [123]. For example, recent work has reverse-engineered algorithms developed inside Transformers, including at the 100 million parameter scale [116, 124, 162]. Again ideally, the algorithm discovered could then be executed as a white-box deterministic program.

ELECTRONIC SUPPLEMENT

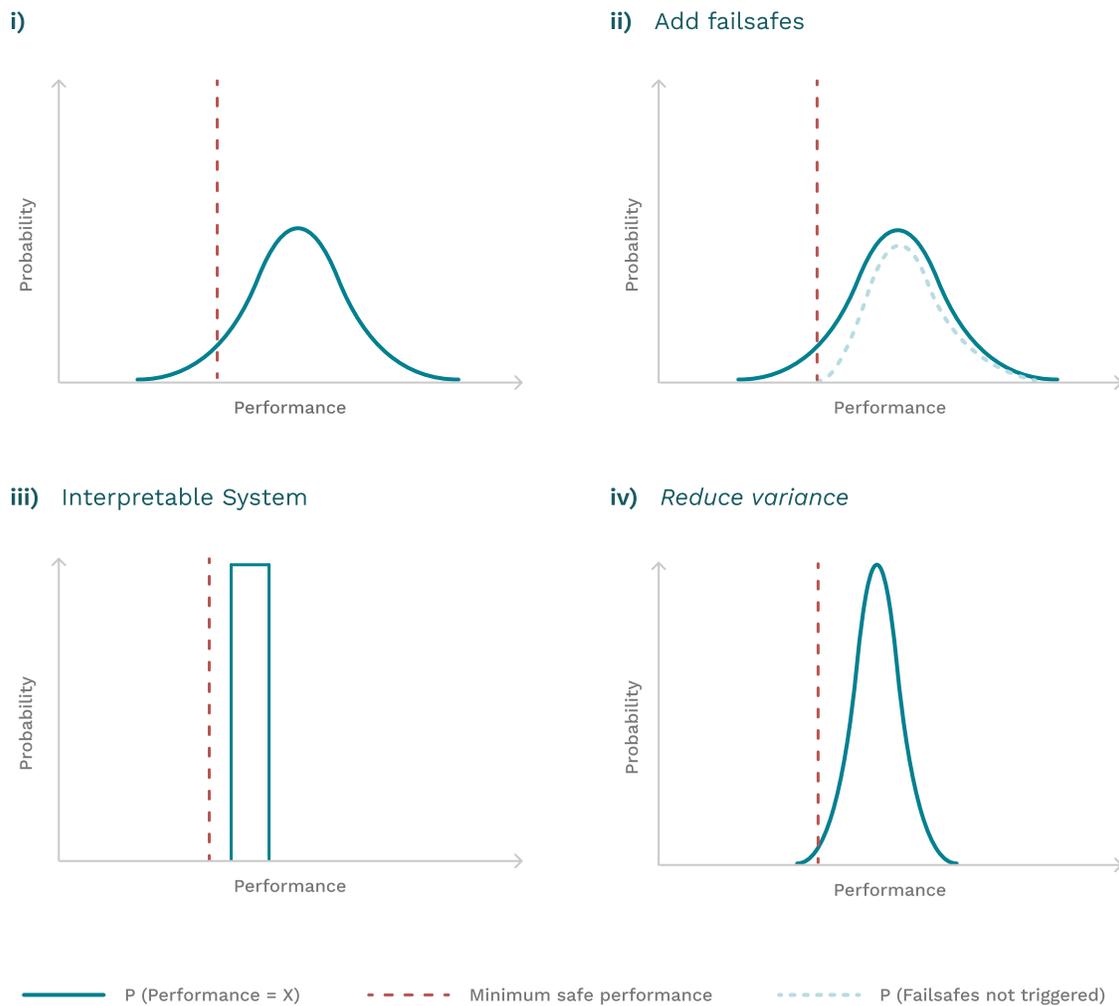

Fig. 14. Assurance is the attempt to reduce the probability of critical failure to an acceptable level. Panel (i) depicts an AI system that broadly performs acceptably, but which has an unacceptable region (to the left of the dotted red line). This might correspond to an AI radiologist misdiagnosing an extremely difficult mammogram, or a car driving on a slick and winding road. The approach in panel (ii) detects poor conditions and refuses to operate, which an be achieved by gating AI with less advanced systems that have well-understood failure modes; panel (iii) depicts a system that has a limited operational range, but that range can be better understood; panel (iv) shows a system designed to have reduced variance in performance, perhaps designed using a system designed to produce "adversarial robustness." Interventions which reduce variance while also reducing average performance can be worthwhile, if they reduce the probability of catastrophic failures.

ELECTRONIC SUPPLEMENT

For tabular data, generalized additive models are an inherently interpretable model class (implemented in the open-source InterpretML library [120]). Similarly, the fledgling subfield of mechanistic interpretability also recently received a dedicated library which helps open the black box of neural networks [117].

Today we have only rudimentary control over high-level properties of trained models like values and goals [54].[8] Many current systems display some form of misaligned behavior [94, 97, 98, 17, 91, 152].

## A.3 Additional Citations: HP#3

*Approaches* The alignment field is under active development, with new perspectives and concepts arising and with a range of different problem decompositions, naming conventions [11] and contested hypotheses [27]. In their data-driven survey, Kirchner et al. discover five clusters in the alignment literature [89]:

(1) *[Empirical] agent alignment* concerns current "agentic" systems, i.e. those trained to perform complex actions in a simulated or real environment, usually with RL. Representative work includes safe RL and reward learning [63, 81], assistance games [146, 170], and adversarial training [174].

(2) *Foundational work* aims to develop strong theoretical frameworks to guide alignment research. Representative work includes agent foundations (formal theories of agency and decision-making) [150, 36], causal incentives (analyzing when a system is in practice being trained for) [88], and natural abstractions [166].

(3) *Tool alignment* concerns nonagentic systems, i.e. classic classification, regression, or generative systems. Representative work includes ML safety [10, 68], the alignment of large language models [128, 86, 67, 87], adversarial training [79, 174], scalable oversight [18] and interpretability research [73, 122].

(4) *AI governance* concerns social and political factors in the transition to advanced AI. (See Section 3.9.)

(5) *Value alignment*, "understanding and extracting human preferences, and designing methods that stop AI systems from acting against these preferences" [89]. Representative work includes [101, 21, 54, 50, 134].

## A.4 Additional Citations: HP#4

### A.4.1 AI to promote access
One great promise of AI is its potential for greater inclusion, for instance of those with disabilities or those who do not speak the dominant language.

*Access for people with disabilities* People with disabilities have traditionally lacked power in industrialized societies [55]. As a result, people with disabilities have often been needlessly disenfranchised by technological progress. Legislation such as the Americans with Disabilities Act of 1990 attempts to address this injustice by legal mandating accessibility. In so doing, such legislation also supports a market for accessibility devices, allowing the cost of R&D to be spread over a much larger customer base.

Technology can both empower and disadvantage people with disabilities: the same speech-to-text technology that powers live captioning of videos on services such as YouTube and Zoom also powers voice assistants—devices which are almost entirely unusable by people who are deaf. In order for all to benefit from the promise of AI, it is critical that AI systems be designed with all users in mind. Promising new developments include brain-computer interfaces to allow people with paralysis to operate computers [121] or, eventually, wheelchairs and powered prostheses [139, 137]. Too often, developers of technology overlook straightforward modifications that could make their systems more accessible. For example, speech-to-text systems have poor accuracy for deaf users out-of-the-box, but accuracy can be dramatically improved if the systems are explicitly trained on the voices of deaf individuals [47].

*Linguistic access* For more than fifty years, English has been the dominant language of science and technology. However, the majority of the world's population do not speak English, and the majority of English speakers are not native speakers. One side effect of this is that "non-English language articles are commonly excluded from published systematic reviews" [138]—and this is true of the present article.

---

[8]This section factors out risks from intentional misuse of AI, and instead asks how to handle the *inherent* risks of intelligent systems, particularly advanced systems with broad capabilities. We discuss misuse risk in sections 3.2 and 3.8.

High-fidelity translation, including the translation of *all* written text, spoken language, and sign language into all other languages, is now well within our capabilities and will dramatically ease linguistic barriers worldwide. Crucially, it will also allow non-English speakers to access large-scale AI systems which use English as the primary interface. While systems such as GPT-4 can translate between English and non-English languages, their performance in English is generally better than in non-English languages due to the availability of training data. The movement to produce language models for non-English languages (along with multilingual models) is thus of enormous importance, given that the overwhelming majority of people cannot use the English-language models which are the current focus of innovation [164].

## A.5 Additional Citations: HP#5

Economic studies of technological impacts on society have a long history, and the analysis of AI fits into this tradition well. The economics of AI is an established subfield within growth theory and the economics of innovation [107, 40], with regular conferences and contributions from top economists [2, 4, 112]. Other approaches for incorporating AI effects into growth theory include [65, 141, 60].

However, the literature has serious limitations. The operational definitions of AI used are often vague and even internally inconsistent. Theoretical studies usually use a broad definition of AI that encompasses most potential use cases, but empirical studies tend to use a narrow definition of AI, often as a synonym for robotics, computer numerical control, robotic process automation, and traditional automation methods [107].

AI is best thought of as a general purpose technology (GPT)—a "single generic technology, recognizable as such... [that] comes to be widely used, to have many uses, and to have many spillover effects" [103, 156]. Past examples of such technologies include electricity, lean production processes, and the internet [103].

However, data on AI tool use are extremely limited. In contrast, data on the use of industrial robots are available at the national level [3], but even these data are mostly limited to summary statistics: as Seamans and Raj highlight, we lack data at the level of individual firms [144]. This precludes in-depth studies on fundamental questions, such as whether robotics and AI complement or substitute labor while simultaneously clearly eliminating jobs. Other classic studies on the impact of AI on automation include [127, 119, 12].

Another approach to ameliorating potential effects on human labor involves requiring companies to hire a certain number of workers per large unit of profit [92].

A natural next step for AI economics is to collect granular data on capabilities and on firm-level and intra-firm utilization. We must also model the labor effects of specific AI technologies in detail, as in [142].

## A.6 Additional Citations: HP#6

*Disciplinary diversity*   Because the Hard Problems are interdisciplinary and often contain social and political subproblems, solving them will require insights from across the sciences and humanities [78, 136]. Many fields have responded to the rise of AI with a new subfield—the economics of AI, the philosophy of AI, and so on.

ELECTRONIC SUPPLEMENT

## A.7 Additional Citations: HP#7

| | Principal citation | Associated legislation | Associated technical agenda |
|---|---|---|---|
| European Commission | [44] | [1] | |
| White House | [16] | [49] | |
| OECD | [154] | | |
| UK | [118] | [159] | |
| China | [43] | [157] | |
| AI4People | [7] | | |
| UNESCO | [160] | | |
| IEEE | [75] | | |
| PAI | [129] | | [130] |
| Asilomar | [77] | | [6] |
| AI Now | [28] | | |
| NeurIPS | [23] | | |
| FAT/ML | [46] | | [135] |
| Montréal | [32] | | [113] |
| OpenAI | [126] | | [125] |
| Meta | [45] | | [5] |
| DeepMind | [33] | | [34] |
| Google | [58] | | [59] |

Table 1. Citations for Figure 10

## A.8 Additional Citations: HP#8

This problem covers foreign policy, the stability of the international order, and potential international governance mechanisms under current and coming AI shocks. (HP#9 focuses instead on domestic governance.)

AI systems have a range of potential applications in militaries and national security [61, 31, 131].

In the middle of the spectrum, conventional warfare is becoming faster and more unpredictable due to the use of AI systems to operate remote platforms, interpret sensor information, and accelerate tactical decision-making, raising the risk of 'flash wars' [48] caused by loss of control, data poisoning issues [105] or destabilizing interactions between AI systems.

AI can, and thus may, change how states use targeted cyber operations and coercion against opponents [53].

Confidence-building measures have been used to stabilize expectations and avoid miscommunication around military technologies, and these measures could play a key role in governing military AI [71, 72].

States have a shared interest in preserving stability, preventing arms races, and preventing proliferation among malicious actors, and can draw on a long history of arms control agreements [143, 108]. Militaries may increasingly come to recognize that technological supremacy does not always equate to improved national security, if it drives one or both sides to play 'technology roulette' [30] by prematurely adopting unreliable or insecure systems.

A classic case study of how technologies can impact international relations deals with the development of armored "dreadnought" battleships, which sparked an arms race in the run-up to World War I [161].

Digital technologies of the past enabled non-state civil society actors to take a much more prominent role participating in and shaping international negotiations [133].

ELECTRONIC SUPPLEMENT

AI may also change the nature of conflict, making wars less predictable and more difficult to limit by augmenting cyber, conventional, and nuclear capabilities [90]. Finally, it may lead to the creation of new players in the geopolitical arena [37].

The AI effect is largely a *structural* risk, in the sense used by Dafoe: "When we think about the risks arising from the combustion engine—such as urban sprawl, blitzkrieg offensive warfare, strategic bombers, and climate change—we see that it is hard to fault any one individual or group for negligence or malign intent... technology can produce social harms, or fail to have its benefits realized, because of a host of structural dynamics. The impacts from technology may be diffuse, uncertain, delayed, and hard to contract over. Existing institutions are often not suited to managing disruption and renegotiating arrangements" [29].

"AI Governance: Overview and Theoretical Lenses" provides an excellent introduction to the current state of AI governance [29].

Although the EU leads in AI regulation, it is struggling to catch up in actual AI research, as well as other areas such as the manufacture of semiconductors [64]. The EU typically employs a risk-based approach that distinguishes between different AI applications and proposes distinct regulations for each, and violators are subject to stiff penalties calculated as percentages of their world-wide revenues. As a result, the EU's regulatory capacity and market size also give it significant influence beyond its borders [19], a phenomenon known as the "Brussels effect" [19, 147]. Seeking to ensure that US tech companies comply with its laws, the EU recently established an office in San Francisco [41].

Although the EU's approach has been praised for its ethical considerations and protection of fundamental rights, the ultimate impact of this regulation is far from clear.

Global AI regulation is still in its early stages, but is developing rapidly: Organization for Economic Co-operation and Development (OECD) members and several other states have agreed to a set of AI Principles, and the G7 has created a Global Partnership on AI. International organizations such as the United Nations Educational, Scientific and Cultural Organization (UNESCO), the Council of Europe, and the OECD have also convened multi-stakeholder groups to draft policy instruments in this area [26]. International standards developed by organizations such as the International Organization for Standardization (ISO) and the Institute of Electrical and Electronics Engineers (IEEE) also play a significant role in AI regulation [148, 25]. In addition, ethical guides and codes serve as important tools for cross-border AI governance, providing guidance for companies and other stakeholders [84, 151].

Some important contributions to the field have dealt with finding consensus on principles and developing global AI governance proposals. Some scholars have suggested the creation of centralized intergovernmental agencies [25] to coordinate global policy responses, while others have proposed the establishment of an International Artificial Intelligence Organization [42] or an international coordinating mechanism under the G20 [80, 26].

## A.9  Additional Citations: HP#10

An autonomous driving championship, Roborace, closed after six seasons owing to lack of interest and funding [8].

Perhaps most people lack the technical or legal knowledge to understand their data, data protection law, and how (or even the fact that) a system is making automated decisions [145].

Landmore provides an example of speculative technologies which could instead enhance democratic participation in [96].

ELECTRONIC SUPPLEMENT

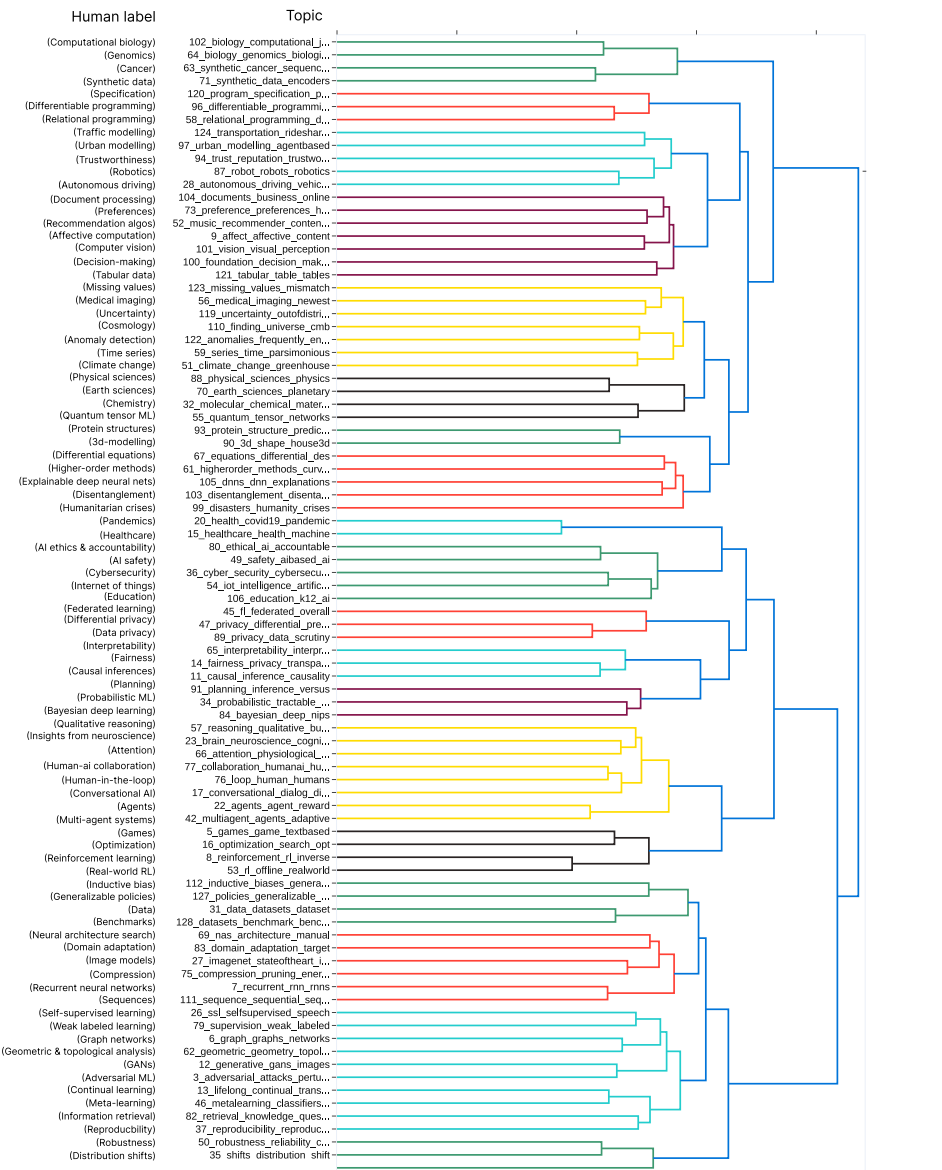

Fig. 15. The topics of presentations at conference workshops between 2017 and 2023, modelled using BERTopic [62]. The conferences covered are AAAI, ICLR, NeurIPS, IJCAI, and ICML. Each topic is named after its three most common words ('Topic'), and was also given a name for readability ('Human label').

## B  Topic modelling of top ML conferences

Investigating our Hard Problems further, we tested a more quantitative approach to surveying AI research as a whole. Figure 15 shows how research areas cluster together based on word co-occurrences in conference workshop text. (Clustering is not a well-defined problem, however, and so each dataset affords many solutions [62].) Our code is available on GitHub here: https://github.com/cufbas/topicModeling; the links used to scrape the text data can be found here: https://zenodo.org/record/7565722. No hyperparameters were altered—Figure 15 represents the first and only clustering run. The only post-processing was to delete spurious topics (those which were overfit or lacked a real common theme).

What could we learn from this? Two possible aims are 1) validating our list of hard problems as a relatively natural categorization of AI research; 2) attempting to discover cross-cutting intellectual movements and schools within AI, which might reveal themselves in the process of naming and organizing workshops (and ultimately new conferences).

The method offers only weak evidence on (1) as the workshops included largely concern AI in a purely technical sense, mostly with regard to relatively abstract capabilities. Figure 15 is thus more of an elaboration of HP#1 and 4.

Regarding (2): movements within AI research are often hard to demarcate, as demonstrated by the overlapping research areas within 'Trustworthiness', 'Interpretability', 'Explainability', 'AI Safety', and 'Algorithmic fairness'. Nevertheless, we think this clustering exercise offers some insight.

ELECTRONIC SUPPLEMENT

ELECTRONIC SUPPLEMENT

ELECTRONIC SUPPLEMENT

ELECTRONIC SUPPLEMENT

ELECTRONIC SUPPLEMENT

ELECTRONIC SUPPLEMENT

ELECTRONIC SUPPLEMENT

ELECTRONIC SUPPLEMENT

ELECTRONIC SUPPLEMENT